\PassOptionsToPackage{hidelinks}{hyperref}
\documentclass{article}
\usepackage[OT1]{fontenc}
\usepackage{newunicodechar}
\newunicodechar{×}{\ensuremath{\times}}

\usepackage{iclr2027_conference,times}

\usepackage{amsmath,amsfonts,bm}

\def\eqref#1{equation~\ref{#1}}

\def\1{\bm{1}}

\DeclareMathAlphabet{\mathsfit}{\encodingdefault}{\sfdefault}{m}{sl}
\SetMathAlphabet{\mathsfit}{bold}{\encodingdefault}{\sfdefault}{bx}{n}

\usepackage{amsmath}
\usepackage{amssymb}
\usepackage{booktabs}
\usepackage{graphicx}
\usepackage{hyperref}
\usepackage{microtype}
\usepackage{placeins}
\usepackage{url}
\usepackage{xcolor}

\title{LeCompass: Task-Level Execution through State-Indexed Interfaces}

\author{%
\normalfont
\begin{tabular}[t]{@{}c@{}}
Kai Shi\textsuperscript{1}, Shiqi Zhao\textsuperscript{1}, Zimo Li\textsuperscript{2}, QU Hairong\textsuperscript{1}, Hongyu Ma\textsuperscript{1},\\
Yongsong Yang\textsuperscript{1}, Jiahao Ai\textsuperscript{1}, Ziran YIN\textsuperscript{3}, Peng Yin\textsuperscript{1}\\[0.5em]
{\small \textsuperscript{1}\,City University of Hong Kong\qquad\textsuperscript{2}\,CyberOrigin}\\
{\small \textsuperscript{3}\,The Hong Kong University of Science and Technology (Guangzhou)}
\end{tabular}%
}

\iclrfinalcopy

\makeatletter
\renewcommand{\@maketitle}{%
  \begingroup
  \centering
  {\LARGE\scshape\@title\par}
  \vspace{0.75em}
  \noindent\@author\par
  \vspace{0.3in minus 0.1in}
  \endgroup
}
\makeatother

\begin{document}

\maketitle
\fancyhead{}
\renewcommand{\headrulewidth}{0pt}

\begin{abstract}
Action-conditioned joint-embedding predictive architectures (JEPAs) offer a foundation for predictive decision-making by modeling the consequences of actions in representation space. Yet evaluating local reachability in latent space does not by itself provide a task-level execution interface: intermediate targets must be generated, local inference has finite budgets, and execution requires spatial feedback and persistent interaction state. We introduce LeCompass, a state-indexed execution framework built around a frozen LeWorldModel (LeWM). A proposer supplies intermediate latent targets, while CompassFlow searches the source space of a learned trajectory prior to obtain effective action candidates under limited predictive-evaluation budgets. Patch-token readouts provide spatial feedback, while observation-based memory supports phase transitions and bounded recovery. State-indexed targets support cross-episode context--goal composition. On 1,000 official TEST full-episode Cube trials, the complete LeCompass execution system achieves 86.9\% native 4-cm reach. Separate matched analyses show that predictive ranking raises success from 4.7\% to 44.9\% on TEST PushT and from 33.8\% to 77.7\% on TEST TwoRoom. These results demonstrate task-level execution with task-dependent roles for latent target proposal, predictive search, and observation-grounded control.
\end{abstract}

\section{Introduction}
\label{sec:introduction}

A central promise of world models is to support deliberation: an agent can predict the consequences of candidate actions and choose among them, rather than relying solely on a direct mapping from observations to actions~\citep{ha2018worldmodels,hafner2019planet}. Joint-embedding predictive architectures (JEPAs) learn predictive latent spaces without reconstructing pixels~\citep{lecun2022path,assran2023ijepa,bardes2024vjepa}. LeWorldModel (LeWM) uses an action-conditioned JEPA to rank CEM-sampled action sequences by the distance between their predicted outcome latents and the target latent~\citep{maes2026leworldmodel}, enabling local planning over roughly 25 frames.

Local reachability alone does not provide task-level execution. A short-horizon predictor can evaluate whether candidate actions approach a specified latent target, but it does not determine how to select intermediate targets without oracle waypoints, generate useful action candidates under limited predictive-evaluation budgets, or verify that predicted progress has been realized during execution. In multi-stage manipulation, execution must also account for contact-dependent transitions and recovery from deviations. We study how explicit predictive planning with a frozen short-horizon world model can be organized into task-level closed-loop execution.

We introduce LeCompass, a task-level execution framework that pairs \emph{state-indexed outer execution} with \emph{time-indexed local imagination}. The outer loop re-indexes task progress from observations, while the frozen LeWM predicts over a short temporal horizon. This separates task progress from prediction time, trajectory generation from predictive evaluation, and predicted reachability from realized execution state.

\begin{figure*}[!ht]
    \centering
    \includegraphics[width=\textwidth]{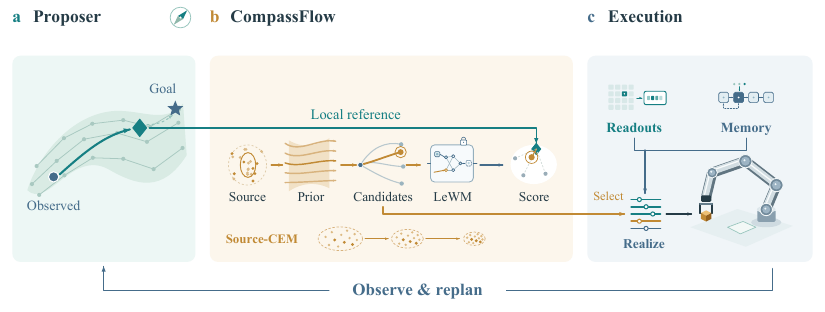}
    \caption{\textbf{Three interfaces from local prediction to task-level execution.}
    (a) Observed history and the final goal yield a local latent reference.
    (b) Target-independent CompassFlow generates action candidates; frozen LeWM
    scores them against the reference, and Source-CEM refines sources over three
    64-sample rounds.
    (c) Cube uses patch-token readouts and observation-based memory for action
    realization, phase transitions, and bounded recovery.}
    \label{fig:local-to-long-horizon}
\end{figure*}

A proposer amortizes demonstration transitions into latent references conditioned on observed history and the final goal, recomputed after each executed segment. CompassFlow pairs a target-independent trajectory prior with frozen-LeWM scoring, so the reference guides candidate selection through predictive evaluation rather than directly conditioning the prior. Patch-token readouts and observation-based memory supply execution geometry, phase transitions, and bounded recovery.

State-indexed proposal raises cross-episode composition success from 2.9\% to 63.7\% on PushT and from 39.1\% to 60.5\% on TwoRoom. At 192 predictive evaluations per plan (approximately 47× fewer than the original action-space CEM budget of 9,000), CompassFlow Source-CEM reaches $52.5\pm2.5$\% on PushT, compared with 16.2\% for action-space CEM at the same budget and $44.9\pm1.1$\% for ranked IID sampling from the same trajectory prior. On TwoRoom, Source-CEM achieves $77.3\pm1.0$\%, compared with 37.2\% for action-space CEM. On the 1,000-episode official Cube panel, the complete system achieves 86.9\% native reach; bounded recovery raises semantic completion under induced contact loss from 5.8\% to 80.4\%.

We contribute:
\begin{itemize}
    \item state-indexed latent references for cross-episode goal composition and reorientation from observed progress;
    \item CompassFlow, a target-independent trajectory prior for low-budget predictive search, with frozen-LeWM scoring and source-coordinate refinement; and
    \item an observation-grounded Cube execution interface with patch-token readouts, temporal state, and bounded recovery.
\end{itemize}

\section{Related Work}
\label{sec:related}

\paragraph{Predictive world models and latent planning.}
Latent world models and JEPAs support representation-space planning~\citep{ha2018worldmodels,hafner2019planet,hansen2024tdmpc2,lecun2022path,assran2023ijepa,bardes2024vjepa}. PLDM and DINO-WM use learned or pretrained visual representations~\citep{sobal2025rewardfree,zhou2025dinowm}, while V-JEPA 2-AC and LeWM add action conditioning for visual control~\citep{assran2025vjepa2,maes2026leworldmodel}. LeCompass embeds a frozen action-conditioned JEPA in a task-level execution system.

\paragraph{Hierarchical targets and temporal abstraction.}
Visual MPC optimizes actions toward visual goals~\citep{finn2017visualforesight}; hierarchical methods add subgoals, latent plans, or high-level goal policies~\citep{nair2020hierarchicalforesight,lynch2019playlmp,hafner2022director,park2023hiql}. SuSIE generates image subgoals~\citep{black2023susie}; $\pi_{0.7}$ conditions a generalist action model on subtask instructions and generated multi-view subgoal images~\citep{physicalintelligence2026pi07}; and HWM learns multi-timescale predictors~\citep{zhang2026hwm}. LeCompass instead uses state-indexed latent references with one frozen short-horizon predictor, requiring neither test-time visual waypoints nor a predictor hierarchy.

\paragraph{Generative trajectory priors and inference-time search.}
Diffusion and flow models generate plans or actions~\citep{janner2022diffuser,chi2023diffusionpolicy,lipman2023flowmatching,zhang2025flowpolicy}; inference-time methods select or refine noise candidates~\citep{ma2025scaling}, including through diffusion mutations~\citep{yang2024diffusiones}. D-MPC combines diffusion-based action proposals and multi-step dynamics for online MPC~\citep{zhou2024diffusionmpc}; GPC ranks and refines generative-policy proposals through predictive world modeling~\citep{qi2025gpc}. DSRL uses reinforcement learning to steer a frozen diffusion policy through its latent-noise space~\citep{wagenmaker2025dsrl}, while Deep Imitative Models combine learned behavior distributions with goal-directed planning~\citep{rhinehart2020deep}. LeFlow generates goal-conditioned latent paths, decodes them through inverse dynamics, and verifies them with a frozen world model~\citep{huang2026leflow}. CompassFlow instead searches source coordinates of a target-independent action-trajectory prior, scoring decoded candidates with frozen LeWM rather than searching raw action space.

\paragraph{Action generation and observation-grounded execution.}
Vision-language-action and world-action models map multimodal context or predicted futures to actions~\citep{kim2024openvla,black2025pi0,ye2026dreamzero,wang2026trustimagination,wang2026worldactionmodels}. INTACT maps latent intent to a largely search-free JEPA action distribution, with optional local CEM~\citep{sun2026intact}. Execution monitoring and behavior trees organize behavior through persistent state, applicability conditions, and recovery~\citep{pettersson2005execution,ogren2022behaviortrees}. LeCompass combines this execution-side organization with test-time candidate generation and frozen-JEPA predictive evaluation.

\section{Problem Formulation}
\label{sec:problem}

We consider goal-conditioned visual control with observation $o_t$, action $a_t$, and goal observation $o_\gamma$. A frozen JEPA encoder provides global planning latents and spatial patch tokens; its action-conditioned predictor estimates candidate outcomes over a supported local horizon $H$:
\begin{equation}
    (z_t,p_t)=E_\theta(o_t),\quad (z_\gamma,p_\gamma)=E_\theta(o_\gamma),\quad
    \hat z_{t+H}=F_\theta(z_t,a_{t:t+H-1}).
    \label{eq:frozen-jepa}
\end{equation}
At replanning round $j$, the controller evaluates at most $B$ candidates, executes a prefix of length $h_j\leq H_j$, and replans from the next observation. Success is defined by a domain-specific task predicate.

\paragraph{State-indexed task progress.}
Observed latent history $\mathbf z_j^{\mathrm{hist}}$, final-goal latent $z_\gamma$, and local span $\kappa_j$ determine a new reference at each replanning round:
\begin{equation}
    z_j^g=P_\psi(\mathbf z_j^{\mathrm{hist}},z_\gamma,\kappa_j),
    \qquad H_j=b\kappa_j,
    \label{eq:state-indexed-target}
\end{equation}
Here $b$ counts raw actions per latent step. Unlike an elapsed-time waypoint schedule, the proposer re-indexes its reference from observed progress.

\paragraph{Target-independent generation and target-dependent evaluation.}
A target-independent trajectory prior maps source coordinates $\xi$ to action candidates, scored against $z_j^g$ using frozen-model predictions:
\begin{equation}
    \mathbf a_j(\xi)=\mathcal T_\phi(\xi;\mathbf z_j^{\mathrm{hist}},\kappa_j),
    \qquad
    \xi_j^\star=\arg\min_{\xi\in\Xi_{B,j}}
    \left\|F_\theta(z_{t_j},\mathbf a_j(\xi))-z_j^g\right\|_2^2.
    \label{eq:source-space-search}
\end{equation}
The evaluated source set $\Xi_{B,j}$ contains at most $B$ coordinates, each requiring one predictive evaluation.

\paragraph{Observation-grounded action realization.}
Predicted proximity to a reference does not establish realized task progress. Spatial readouts and persistent memory therefore govern action realization:
\begin{equation}
    r_t=R_\rho(p_t,p_\gamma),\quad m_t=U_m(m_{t-1},r_t,e_t),\quad
    \tilde a_t=\mathcal{X}(a_t^{\mathrm{plan}};r_t,m_t).
    \label{eq:execution-interface}
\end{equation}
Readouts $r_t$ expose execution geometry, using $p_\gamma$ only for goal-relative outputs. Memory $m_t$ accumulates observation-derived evidence $e_t$, such as contact; $\mathcal X$ combines planned actions, readouts, and memory into the environment action $\tilde a_t$.

The proposer $P_\psi$, prior $\mathcal T_\phi$, and readouts $R_\rho$ are learned with the JEPA frozen; the memory update $U_m$ and action realization $\mathcal X$ are non-learned (Sec.~\ref{sec:execution-memory}).

\section{LeCompass}
\label{sec:method}

\begin{figure*}[!ht]
    \centering
    \includegraphics[width=\textwidth]{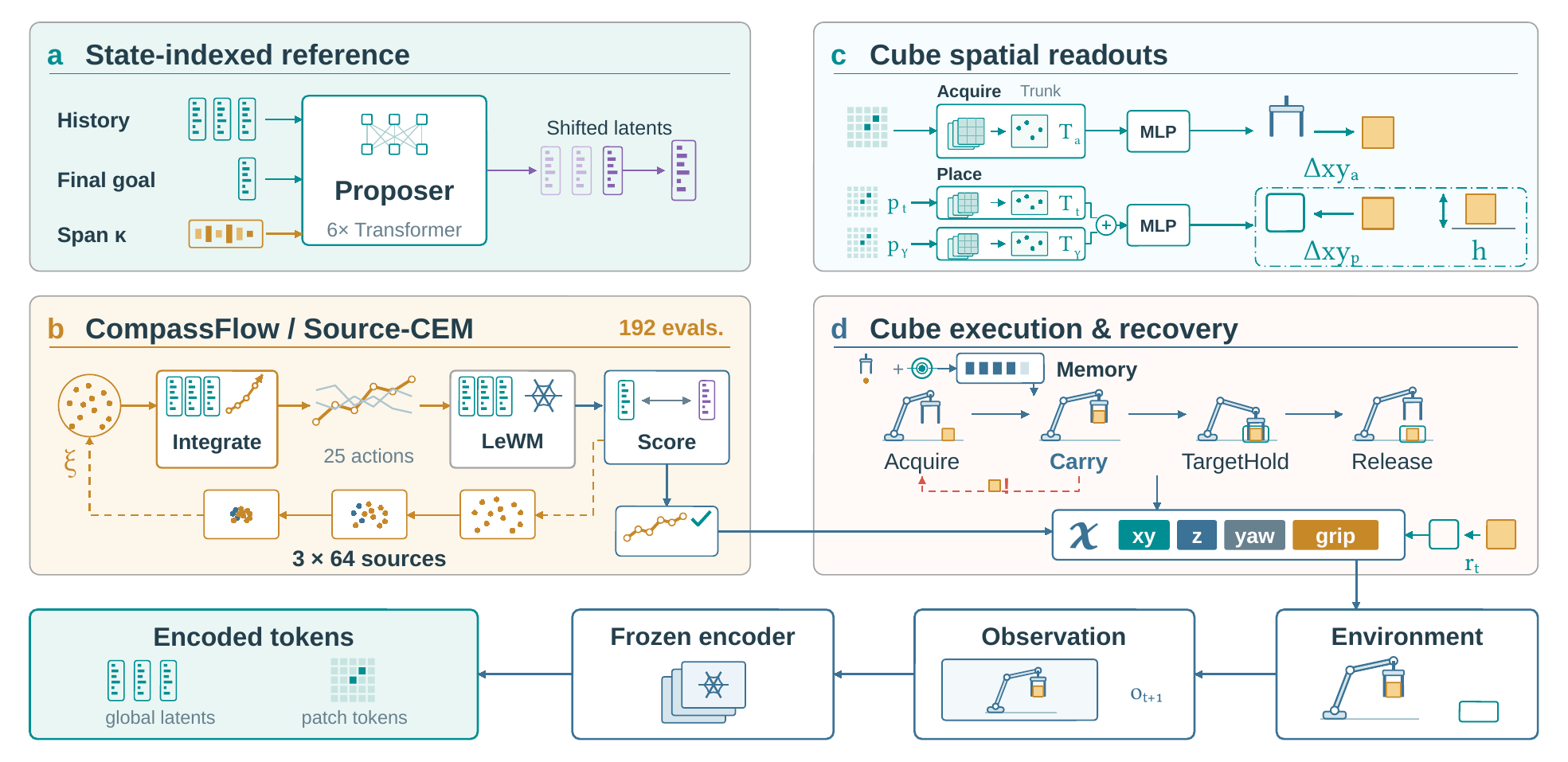}
    \caption{\textbf{LeCompass planning and execution interfaces.}
    \textbf{(a)} History, final goal, and span $\kappa$ condition the proposer; its latest inverse-normalized prediction supplies the scoring reference.
    \textbf{(b)} CompassFlow decodes 25-action candidates with eight Euler steps. Source-CEM evaluates three 64-source rounds, using frozen LeWM to score predicted outcomes. The reference enters scoring, not the trajectory prior.
    \textbf{(c)} Cube acquisition reads current patch tokens; placement also uses cached final-goal tokens to estimate displacement and block height.
    \textbf{(d)} Observation-based memory confirms phases and permits bounded recovery from contact loss in \textsc{Carry}. The placement controller assigns action channels, after which grip-floor and release actuators may override the gripper command. Environment observations update the encoded history and patch tokens.}
    \label{fig:lecompass-architecture}
\end{figure*}

Figure~\ref{fig:lecompass-architecture} summarizes the closed loop of target proposal, trajectory search, and action realization.

\subsection{State-Indexed Latent Proposal}
\label{sec:proposer}

A frozen short-horizon predictor evaluates local reachability but cannot direct successive local predictions. The proposer amortizes directed transitions in demonstration latents into history- and goal-conditioned local references. It maps the latest three normalized latents, a later-goal latent, and span $\kappa$ to a shifted three-latent sequence. The conditioning goal $z_g$ is sampled from the context trajectory during training and set to $z_\gamma$ at evaluation. With latent dimension $D=192$, normalization $\nu$, latent stride $\Delta$, and $H=b\kappa$, training minimizes
\begin{equation}
    \mathcal L_{\mathrm{prop}}=\mathbb E\!\left[
    \frac{1}{3D}\sum_{q\in\{-2,-1,0\}}
    \left\|[P_\psi^{\mathrm{seq}}(\bar{\mathbf z}_t^{\mathrm{hist}},
    \nu(z_g),\kappa)]_q-\nu(z_{t+q\Delta+H})\right\|_2^2\right].
    \label{eq:proposer-loss}
\end{equation}
The inverse-normalized latest prediction supplies $z_j^g$, a synthesized directional reference for candidate scoring rather than a retrieved demonstration frame or a physical waypoint that must be reached exactly.

Recomputing references from observed history supports cross-episode context--goal composition and reorientation after deviations. The proposer and trajectory prior are trained across supported spans; all reported deployments use $\kappa=5$. Appendix~\ref{sec:appendix-proposer} specifies sampling, normalization, and architecture.

\subsection{CompassFlow: Prior-Guided Trajectory Search}
\label{sec:compassflow}

A synthesized reference may lack a paired expert trajectory. CompassFlow learns a state-conditioned, target-independent prior over demonstrated action chunks; references guide test-time selection through frozen-LeWM scoring.

For normalized expert action chunk $\bar{\mathbf a}$, source noise $\epsilon\sim\mathcal N(0,I)$, and interpolation time $\tau\sim\mathcal U[0,1]$, define $x_\tau=(1-\tau)\epsilon+\tau\bar{\mathbf a}$. The rectified-flow velocity field minimizes
\begin{equation}
    \mathcal L_{\mathrm{flow}}=\mathbb E\!\left[
    \underset{M_\kappa=1}{\operatorname{mean}}
    \left(v_\phi(x_\tau,\tau;\bar{\mathbf z}^{\mathrm{hist}},\kappa)
    -(\bar{\mathbf a}-\epsilon)\right)^2\right],
    \label{eq:flow-loss}
\end{equation}
The mask $M_\kappa$ marks the valid $b\kappa$ action positions. The field conditions on observed history and span, not $z_j^g$. Integrating $v_\phi$ defines $\mathcal T_\phi$, implemented with eight Euler steps.

At inference, Source-CEM~\citep{rubinstein1999crossentropy} scores candidates using Eq.~\ref{eq:source-space-search}, allocating a default budget of 192 predictive evaluations across three 64-candidate rounds. It fits a diagonal Gaussian to low-cost source coordinates while retaining base-distribution samples, then selects the lowest-cost trajectory across all rounds. Appendix~\ref{sec:appendix-flow} specifies masking and action normalization; Appendix~\ref{app:source-search} specifies source-bank construction, source constraints, ranked IID sampling, and the complete Source-CEM update.

\subsection{Spatial Readout and Execution Memory}
\label{sec:execution-memory}

Patch-token readouts estimate current geometry; observation-supported memory accumulates confirmed execution conditions across replanning. Together they govern action-channel assignments and overrides, actuator constraints, phase transitions, and prefix invalidation when supporting conditions fail. Memory is non-learned; its structure and task-specific parameters are determined from development rollouts and fixed before evaluation.

\paragraph{Cube instantiation.}
Acquisition uses current patch tokens; placement additionally uses cached final-goal tokens:
\[
    r_t^{\mathrm{acq}}=R_{\rho_a}(p_t),\qquad
    r_t^{\mathrm{place}}=R_{\rho_p}(p_t,p_\gamma).
\]
These goal tokens are independent of the intermediate ranking reference. Readouts are adapted on development data using simulator-state labels, with the encoder frozen, then fixed for evaluation.

After trajectory selection, readouts guide lateral control; remaining planner commands follow phase-specific pass-through and actuator constraints. During pre-pick, vertical and yaw commands pass through, while the observation-based grip floor may override the gripper command. Accumulated contact and placement evidence confirms phases, maintains gripper constraints during carry, and gates release. When recovery is enabled, detecting sustained contact loss in \textsc{Carry} prompts the controller to discard the unexecuted prefix and initiate bounded reacquisition, preserving observed latent history and outer-loop task progress.

The readout targets, evidence predicates, temporal thresholds, channel assignments, and recovery budgets are task-specific. Appendices~\ref{app:cube-execution} and~\ref{app:readout-adaptation} specify the complete Cube execution program and readout adaptation.

\section{Experimental Setup}
\label{sec:setup}

\paragraph{Environments and protocols.}
We evaluate PushT planar object pushing, collision-aware TwoRoom navigation, and full-episode Cube pick-and-place~\citep{maes_lld2026swm,maes2026leworldmodel}. T1 tests nominal execution; T2 pairs contexts and goals from different episodes; T3 perturbs execution while the controller observes and replans. Cube T3 forces gripper opening after stable carry. Task-level and primary search comparisons use official TEST panels; additional Cube component and bare-planner analyses share a 500-task panel. Learned LeCompass components use non-TEST training data; LeWM remains frozen.

\paragraph{Comparisons.}
Target-interface comparisons use the original action-space CEM (9,000 predictive evaluations per plan) with direct final-goal targets, elapsed-time Fixed-10/25 demonstration waypoints, or state-indexed proposals. Search comparisons evaluate action-space CEM, ranked flow sampling, and CompassFlow Source-CEM at 192 predictive evaluations per plan; separate attribution studies vary the scoring target or predictive ranking. The PushT comparison matches proposer refresh cadence, warm-start behavior, observation history, and execution settings. Same-prior ranked IID sampling tests the additional benefit of adaptive refinement. We additionally evaluate all three search methods at 48, 96, 192, 384, and 768 predictive evaluations per plan on PushT and TwoRoom. Cube Recovery ON/OFF changes only the bounded recovery bundle, retaining the same readouts, monitoring, and perturbation rule.

\paragraph{Metrics and reporting.}
PushT uses joint object-and-tool success; TwoRoom requires target-region entry; Cube reports native 4-cm reach and post-release semantic completion. Paired success comparisons use exact McNemar tests; paired-bootstrap confidence intervals quantify uncertainty in search-performance differences and runtime speedups. Evaluation protocols and implementation details are provided in Appendices~\ref{sec:appendix-protocol} and~\ref{app:revision-evidence}.

\section{Results}
\label{sec:results}

\subsection{State-Indexed Target Composition}
\label{sec:results-proposer}

Table~\ref{tab:proposer-results}(a) tests cross-episode context--goal composition without an episode-specific waypoint schedule. Despite training on within-trajectory pairs, state-indexed proposal raises success from 2.9\% to 63.7\% on PushT and from 39.1\% to 60.5\% on TwoRoom. With a test-time demonstration schedule (Table~\ref{tab:proposer-results}(b)), it is comparable to Fixed-25 on PushT T1 and stronger under T3 perturbation (29.0\% versus 20.2\%; exact McNemar $p=7.9\times10^{-4}$), consistent with re-indexing targets from observed progress. Fixed-25 remains stronger on TwoRoom T1 and T3; Sec.~\ref{sec:analysis-resolvability} examines the related frozen-latent geometry.

\begin{table}[!ht]
    \centering
    \small
    \setlength{\tabcolsep}{5pt}
    \caption{\textbf{State-indexed target composition.} Success (\%) on official TEST panels with 9,000 predictive evaluations per plan. (a) Cross-episode T2 without an episode-specific waypoint schedule. (b) T1/T3 with test-time demonstration schedules; Fixed-$k$ advances through demonstration waypoints every $k$ raw steps. -- denotes an unrun arm. State-indexed PushT T1 is mean $\pm$ sample SD over three proposer checkpoints; all other entries are single fixed runs (Appendix~\ref{sec:appendix-protocol}).}
    \label{tab:proposer-results}

    \vspace{2pt}
    \textbf{(a) No episode-specific waypoint schedule: cross-episode T2}

    \vspace{2pt}
    \begin{tabular}{@{}lrrr@{}}
        \toprule
        Domain & Context--goal pairs & Direct final goal & State-indexed target \\
        \midrule
        PushT   & 1,000 & 2.9  & \textbf{63.7} \\
        TwoRoom & 1,000 & 39.1 & \textbf{60.5} \\
        \bottomrule
    \end{tabular}

    \vspace{6pt}
    \textbf{(b) Episode-specific demonstration schedule available}

    \vspace{2pt}
    \begin{tabular}{@{}llrrrrr@{}}
        \toprule
        Domain & Protocol & $N$ & Direct & Fixed-10 & Fixed-25 & State-indexed \\
        \midrule
        PushT
            & T1 & 500 & 5.2 & 57.0 & \textbf{59.2} & 58.1$\pm$1.0 \\
            & T3 & 500 & 4.2 & --   & 20.2 & \textbf{29.0} \\
        \midrule
        TwoRoom
            & T1 & 500 & 14.0 & 29.0 & \textbf{40.6} & 31.2 \\
            & T3 & 500 & --   & --   & \textbf{32.6} & 28.4 \\
        \bottomrule
    \end{tabular}
\end{table}

\subsection{Finite-Budget Prior-Guided Search}
\label{sec:results-executor}

At 192 predictive evaluations per plan, CompassFlow Source-CEM reaches $52.5\pm2.5$\% success on PushT and $77.3\pm1.0$\% on TwoRoom, versus 16.2\% and 37.2\% for action-space CEM (Table~\ref{tab:executor-results}). Same-prior ranked IID sampling reaches $44.9\pm1.1$\% and $77.3\pm0.4$\%, respectively. Within Source-CEM, replacing the final-goal scoring target with the proposed reference improves mean success by 38.6 percentage points on PushT and 15.3 points on TwoRoom. Matched comparisons with predictively ranked IID sampling show that adaptive source refinement adds 6.2--9.4 points across the three PushT checkpoints but no mean gain on TwoRoom (Appendix~\ref{sec:appendix-executor-attribution}). With shared Flow initialization and matched CEM settings on PushT, source-space refinement achieves 52.93\% versus 49.33\% for action-space refinement, a mean improvement of 3.6 percentage points (Appendix~\ref{app:revision-coordinate}). For reference, the earlier action-space CEM configuration with the proposed target achieves \(58.1\pm1.0\)\% on PushT T1 at 9,000 predictive evaluations per plan (Table \ref{tab:proposer-results}).

\begin{table}[!ht]
    \centering
    \small
    \setlength{\tabcolsep}{3pt}
    \caption{\textbf{Finite-budget local inference.} Success (\%) at 192 predictive evaluations per plan on official TEST PushT (500 episodes) and strict TEST TwoRoom (500 windows). $h$ is the nominal maximum execution prefix in raw actions. Flow-based methods report mean $\pm$ sample SD over three field checkpoints; action-space CEM uses single fixed runs. Final-goal rows retain the existing target-attribution runs.}
    \label{tab:executor-results}
    \begin{tabular}{@{}llllr@{}}
        \toprule
        Domain & Search & Target & $h$ & Success (\%) \\
        \midrule
        PushT & Action-space CEM & Proposed latent & 5 & 16.2 \\
              & Flow ranked IID & Proposed latent & 5 & $44.9\pm1.1$ \\
              & CompassFlow Source-CEM & Proposed latent & 5 & \textbf{52.5$\pm$2.5} \\
              & CompassFlow Source-CEM & Final-goal latent & 5 & $13.9\pm1.2$ \\
        \midrule
        TwoRoom & Action-space CEM & Proposed latent & 25 & 37.2 \\
                & Flow ranked IID & Proposed latent & 25 & $77.3\pm0.4$ \\
                & CompassFlow Source-CEM & Proposed latent & 25 & \textbf{77.3$\pm$1.0} \\
                & CompassFlow Source-CEM & Final-goal latent & 25 & $62.0\pm2.0$ \\
        \bottomrule
    \end{tabular}
\end{table}

\paragraph{Budget sensitivity.}
A fixed-checkpoint sweep evaluates all three methods over budgets from 48 to 768 predictive evaluations per plan. On PushT, Source-CEM reaches 38.4--54.8\%, versus 12.2--18.2\% for action-CEM. On TwoRoom, Source-CEM success rises from 62.2\% to 84.4\% as the budget increases from 48 to 768, closely tracking ranked IID sampling (64.4\% to 84.2\%). Source-CEM consistently outperforms action-space CEM across the evaluated budgets, while its additional benefit over same-prior IID sampling depends on the task and budget (Appendix~\ref{app:revision-budget}).

\subsection{Full-Episode Cube Execution and Recovery}
\label{sec:results-cube}

On the common 1,000-episode official TEST panel, LeCompass achieves 86.9\% native reach and 86.0\% semantic completion, versus 3.7\% and 4.3\% native reach for Direct and Fixed-25 (Table~\ref{tab:cube-results}(a)). These baselines lack the same goal-aware execution program, so the gap does not isolate predictive planning. Under the matched T3 forced-opening intervention, enabling the bounded recovery bundle raises full-panel semantic completion from 5.8\% to 80.4\%; both arms retain execution monitoring (Table~\ref{tab:cube-results}(b)).

\begin{figure*}[!ht]
    \centering
    \includegraphics[width=\textwidth,trim=0 10pt 0 0,clip]{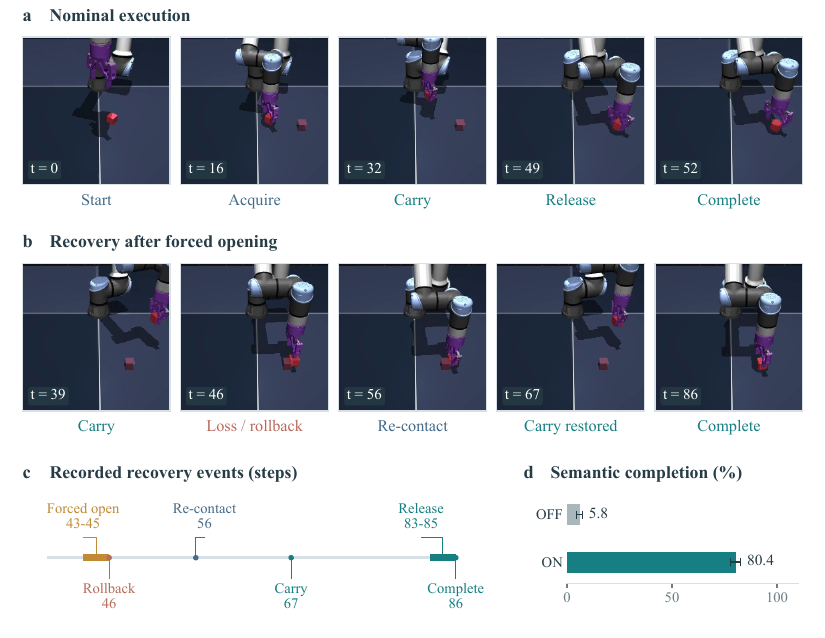}
    \caption{\textbf{Observation-grounded Cube execution and bounded recovery.}
    (a,b) Nominal execution and forced-opening recovery in different official TEST
    episodes; frames are shown at nonuniform steps.
    (c) Forced opening at $t=43$--$45$ leads to rollback at $t=46$, discarding
    three unexecuted actions.
    (d) Matched full-panel T3 semantic completion (1,000 episodes): $5.8\%$ OFF
    versus $80.4\%$ ON; whiskers show 95\% Wilson intervals.}
    \label{fig:cube-execution}
\end{figure*}

Figure~\ref{fig:cube-execution} illustrates nominal execution and bounded recovery.

\begin{table}[!ht]
    \centering
    \small
    \setlength{\tabcolsep}{5pt}
    \caption{\textbf{Full-episode Cube execution and bounded recovery.} (a) T1 native 4-cm reach, semantic completion, and predictive budget on 1,000 official TEST episodes. Native reach includes frame zero; completion requires completed release and post-third-opening containment. $B/h$ is nominal predictive evaluations per environment step; -- denotes not evaluated. (b) Matched T3 Recovery ON/OFF under the same perturbation, readouts, and monitoring; only bounded recovery changes. Further diagnostics are in Appendix~\ref{sec:appendix-cube-breakdowns}.}
    \label{tab:cube-results}

    \vspace{2pt}
    \textbf{(a) T1: full-episode execution and predictive budget}

    \vspace{2pt}
    \begin{tabular}{@{}lrrrr@{}}
        \toprule
        Method & \shortstack{Native reach\\(\%) $\uparrow$} & \shortstack{Semantic completion\\(\%) $\uparrow$} & \shortstack{Evals/\\plan $\downarrow$} & \shortstack{Nominal\\$B/h$ $\downarrow$} \\
        \midrule
        Direct final-goal CEM & 3.7 & -- & 9,000 & 360.0 \\
        Fixed-25 demonstration-oracle CEM & 4.3 & -- & 9,000 & 360.0 \\
        LeCompass & 86.9 & 86.0 & 192 & 38.4 \\
        \bottomrule
    \end{tabular}

    \vspace{6pt}
    \textbf{(b) T3: full-panel bounded-recovery intervention}

    \vspace{2pt}
    \begin{tabular}{@{}lrr@{}}
        \toprule
        Outcome & Recovery ON & Recovery OFF \\
        \midrule
        Native reach & 837/1000 (83.7\%) & 397/1000 (39.7\%) \\
        Semantic completion & \textbf{804/1000 (80.4\%)} & 58/1000 (5.8\%) \\
        \bottomrule
    \end{tabular}

\end{table}

\paragraph{Predictive ranking and execution components.}
On a shared panel of 500 Cube tasks, predictive ranking raises bare IID-planner native reach from 6.6\% to 25.4\%, a gain of 18.8 percentage points (95\% CI $[15.0,22.6]$; Table~\ref{tab:component500}(a)). With Source-CEM, planner-only reaches 27.4\%, contact-based memory-only 32.2\%, and the full interface 87.8\% (Table~\ref{tab:component500}(b)). The latter gain combines learned spatial feedback with the execution logic that uses it. Appendix \ref{sec:appendix-cube-simple-control} reports a shared-stack scripted reference with additional proprioceptive height.

\begin{table}[!ht]
    \centering\small
    \setlength{\tabcolsep}{3pt}
    \caption{\textbf{Cube planning and execution ablations.} Native reach on the same 500 tasks, including the initial state. (a) Both bare planners decode 192 IID proposals; only the ranked arm scores them with LeWM. (b) All three arms use Source-CEM192 with different execution components. Protocols: Appendices~\ref{app:revision-cube} and~\ref{app:cube-bare-ranking}.}
    \label{tab:component500}
    \begin{tabular}{@{}lllrr@{}}
    \toprule
    Configuration & Readouts & Execution control & $n$/500 & Native (\%) \\
    \midrule
    \multicolumn{5}{@{}l}{\textbf{(a) Predictive ranking without the execution interface}} \\
    IID, first candidate & None & None & 33 & 6.6 \\
    IID, WM-ranked & None & None & 127 & 25.4 \\
    \midrule
    \multicolumn{5}{@{}l}{\textbf{(b) Source-CEM execution components}} \\
    LeCompass planner-only & None & None & 137 & 27.4 \\
    Memory-only & None & Grip protection & 161 & 32.2 \\
    Full & Acquisition + placement & Full program & 439 & 87.8 \\
    \bottomrule
    \end{tabular}
\end{table}

\begin{figure*}[!ht]
    \centering
    \includegraphics[width=\textwidth]{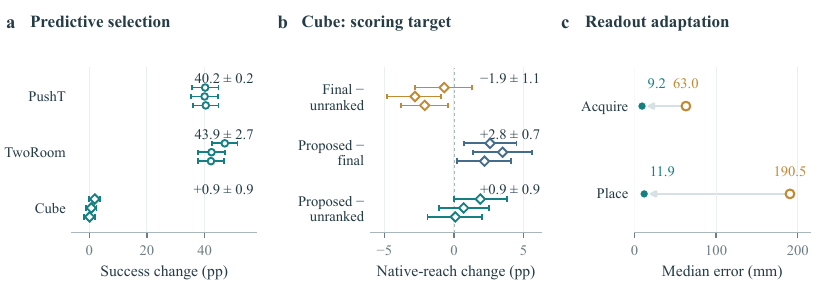}
    \caption{\textbf{Predictive selection and spatial readout evidence.}
    \textbf{(a)} Proposed-target IID best-of-192 selection versus matched unranked execution on official TEST PushT, strict TEST TwoRoom, and Cube. Points represent three field checkpoints for PushT/TwoRoom or three source seeds for Cube; whiskers are within-repeat paired 95\% intervals, and labels give mean $\pm$ sample SD.
    \textbf{(b)} Scoring-target contrasts in the same Cube IID suite, separate from the headline run; the net proposed-minus-unranked effect remains unresolved.
    \textbf{(c)} Initial (open amber) and adapted (filled teal) median XY readout errors on identical controller-induced frames: 5,338 for acquisition and 5,580 for placement. Separate closed-loop validation improves confirmed picks from $47/113$ to $93/113$ and, with acquisition fixed, placement native reach from $11/93$ to $92/93$ on the same 93 confirmed-pick episodes.}
    \label{fig:attribution-insight}
\end{figure*}

\subsection{End-to-End Computational Cost}
Full LeCompass uses 192 predictive evaluations per plan versus 9,000 for Direct and Fixed-25, an approximately 47-fold reduction. Mean rollout wall times yield $1.16\times$, $7.94\times$, and $2.67\times$ speedups over Direct on PushT, TwoRoom, and Cube, respectively (Table~\ref{tab:runtime50}). Replanning and stopping differ across systems; Cube's savings include earlier program termination.

\begin{table}[!ht]
    \centering\small
    \setlength{\tabcolsep}{4pt}
    \caption{\textbf{End-to-end rollout cost.} Mean rollout time (seconds/task) over 50 tasks per domain, excluding model loading and task setup. Parentheses give predictive evaluations per plan.}
    \label{tab:runtime50}
    \begin{tabular}{@{}lrrrr@{}}
    \toprule
    Domain & Direct (9000) & Fixed-25 (9000) & LeCompass (192) & Speedup vs. Direct \\
    \midrule
    PushT & 4.267 & 10.495 & 3.673 & $1.16\times$ \\
    TwoRoom & 3.563 & 3.366 & 0.449 & $7.94\times$ \\
    Cube & 43.763 & 42.318 & 16.383 & $2.67\times$ \\
    \bottomrule
    \end{tabular}
\end{table}

Separate 192-budget profiling yields 69--70\,ms per Source-CEM decision versus 58--59\,ms for action-space CEM at the same budget, including about 10\,ms for Source-CEM's three sequential Flow decodes. Module costs, execution counts, and cross-time repeats are in Appendix~\ref{app:revision-runtime}.

\section{Analysis}
\label{sec:analysis}

\subsection{How Does Target Geometry Differ Across Domains?}
\label{sec:analysis-resolvability}

Let $E$ be median proposer error, $G$ the 95th-percentile global-pair distance, and $A$ the median one-latent-step transition distance. The ratio $E/(G-A)$ is 0.137 on PushT, 0.227 on Cube, and 2.924 on TwoRoom. TwoRoom's narrow local-to-global separation relative to proposer error is consistent with Fixed-25's advantage when an episode-specific demonstration schedule is available (Table~\ref{tab:proposer-results}). This checkpoint-conditioned distance diagnostic does not measure dynamical reachability; Appendix~\ref{sec:appendix-resolvability} reports the full scales and uncertainty intervals.

\subsection{What Does Predictive Search Contribute?}
\label{sec:analysis-attribution}

Predictive selection improves bare planning, but its incremental contribution depends on the execution interface. With the encoder and trajectory prior held fixed, proposed-target IID best-of-192 selection improves success over unranked execution by $40.2\pm0.2$ percentage points on PushT and $43.9\pm2.7$ points on cadence-25 TwoRoom (Fig.~\ref{fig:attribution-insight}(a)). On Cube, predictive ranking raises native reach from 6.6\% to 25.4\% on the 500-task panel without spatial readouts or the execution program (Table~\ref{tab:component500}(a)). In a separate 1,000-episode suite with the complete execution interface, unranked execution already reaches $86.4\pm1.0$\%, and proposed-target ranking adds an unresolved $0.9\pm0.9$ percentage points (Fig.~\ref{fig:attribution-insight}(b); Appendix~\ref{sec:appendix-executor-attribution}).

\subsection{Why Does Execution Need Observation-Grounded State?}
\label{sec:analysis-execution-state}

The tested projected-CLS MLP decodes end-effector XY well ($R^2=0.957$), but block-minus-effector XY has 90.7\,mm median error and $R^2=-0.128$ on 14,656 validation frames (Appendix~\ref{app:readout-representation}); the spatially corresponding patch token ranks first in all seven non-null swaps. With the encoder and predictor frozen, readout adaptation reduces median acquisition and placement errors from 63.0 to 9.2 mm and from 190.5 to 11.9 mm on shared controller-induced states (Fig.~\ref{fig:attribution-insight}(c)). Separate validation rollouts show corresponding pick and placement gains (Appendix~\ref{app:readout-adaptation}). Readouts provide current geometry, while memory confirms execution conditions. Table~\ref{tab:component500} reports the benefit of the complete spatial execution interface. LeWM scores five-channel candidates before execution-time channel assignments (Appendix~\ref{sec:appendix-cube-breakdowns}).

\section{Limitations}
\label{sec:limitations}

Evaluation covers three simulation benchmarks and released LeWM checkpoints, not physical robots. The complete observation-grounded execution interface is instantiated only on Cube. Transfer beyond Cube may require new geometric supervision for the learned readouts and adaptation of the task-specific execution logic, including its memory, action-channel assignments, and recovery conditions.

Spatial readouts are used by the execution interface but are not integrated into predictive candidate scoring. Candidate actions are evaluated before execution-time corrections; the planner does not re-evaluate the corrected actions. Search-refinement gains remain task-dependent, and predictive ranking provides limited additional benefit with the complete Cube execution interface. Recovery is evaluated under a single induced contact-loss scenario; robustness to natural contact loss and repeated failures remains untested.

\section{Conclusion}
\label{sec:conclusion}

LeCompass organizes frozen-JEPA local prediction into task-level execution through state-indexed latent references, prior-guided predictive search under limited budgets, and observation-grounded control. State-indexed targets support cross-episode composition, and predictive selection improves PushT, TwoRoom, and bare Cube planning. Spatial readouts and execution memory support full-task Cube execution, while bounded recovery improves outcomes after induced contact loss. Together, these results show how explicit predict-and-score decisions from a frozen short-horizon model can sustain task-level closed-loop behavior: observations continually update the planning reference, predicted consequences guide action selection, and execution feedback governs task progression.

\section*{AI Use Statement}

We used AI tools to assist with manuscript drafting and language refinement, and improvements to the organization and consistency of the main text, figures, tables, and appendices. These tools also supported literature retrieval and discovery, including identifying relevant work and suggesting search directions. For research ideation, we used them to discuss methodological choices and refine experimental design. The authors reviewed and revised the AI-assisted content, made the final methodological and editorial decisions, and take full responsibility for the text, references, methods, experiments, analyses, and claims presented in this work.

\bibliography{iclr2027_conference}
\bibliographystyle{iclr2027_conference}

\appendix
\suppressfloats[t]

\section{Experimental Protocols}
\label{sec:appendix-protocol}

\subsection{Evaluation panels and data separation}

Tables~\ref{tab:appendix-target-protocols} and~\ref{tab:appendix-search-protocols} distinguish target-interface, search, and attribution panels. T1 tests nominal execution; T2 pairs contexts and goals from different episodes; T3 perturbs realized execution. Proposer training uses within-trajectory context--goal pairs, whereas T2 uses different TEST episodes excluded from downstream training. PushT T2 uses 1,000 unique episodes in each role, with disjoint role sets and each episode used once.

\begin{table}[!ht]
    \centering
    \scriptsize
    \setlength{\tabcolsep}{2pt}
    \caption{\textbf{Target-interface and full-episode Cube protocols.} $N$ counts episodes for PushT/Cube, strict windows for TwoRoom, and context--goal pairs for T2. Evals. denotes predictive evaluations per plan. Repeats reuse the same panel; TEST denotes the official split. Checkpoints and randomness are specified in Appendix~\ref{app:protocol-randomness}.}
    \label{tab:appendix-target-protocols}
    \begin{tabular}{@{}lllrlllp{0.31\textwidth}@{}}
        \toprule
        Domain & Study & Panel & $N$ & Evals. & $H$ & $h$ & Learned configuration / replication \\
        \midrule
        PushT & Proposer T1 & TEST & 500 & 9,000 & 25 & 25 & 3 proposer training seeds \\
        PushT & Proposer T2 & TEST pairs & 1,000 & 9,000 & 25 & 25 & Validation-selected median run \\
        PushT & Proposer T3 & TEST & 500 & 9,000 & 25 & 25 & Validation-selected median run \\
        \midrule
        TwoRoom & Proposer T1 & Strict TEST & 500 & 9,000 & 25 & 25 & Proposer training seed 0 \\
        TwoRoom & Proposer T2 & Strict TEST pairs & 1,000 & 9,000 & 25 & 25 & Proposer training seed 0 \\
        TwoRoom & Proposer T3 & Strict TEST & 500 & 9,000 & 25 & 25 & Proposer training seed 0 \\
        \midrule
        Cube & LeCompass T1 & TEST & 1,000 & 192 & 25 & 5 & Fixed learned components \\
        Cube & Direct / Fixed-25 T1 & TEST & 1,000 & 9,000 & 25 & 25 & No downstream LeCompass component \\
        Cube & LeCompass T3 & TEST & 1,000 & 192 & 25 & 5 & Same fixed learned components \\
        \bottomrule
    \end{tabular}
\end{table}

\begin{table}[!ht]
    \centering
    \scriptsize
    \setlength{\tabcolsep}{2pt}
    \caption{\textbf{Search and attribution protocols.} $N$ counts episodes for PushT/Cube and strict evaluation windows for TwoRoom. Evals. denotes predictive evaluations per plan; $H/h$ gives the prediction horizon and nominal execution prefix in raw actions. $0/192$ denotes unranked execution versus IID best-of-192 predictive selection. Repeated runs reuse the same panel. Checkpoints and source-sampling seeds are specified in Appendix~\ref{app:protocol-randomness}.}
    \label{tab:appendix-search-protocols}
    \begin{tabular}{@{}lllrccp{0.27\textwidth}@{}}
        \toprule
        Domain & Study & Panel & $N$ & Evals. & $H/h$ & Model / sampling repeats \\
        \midrule
        PushT & Action-space CEM & TEST & 500 & 192 & 25/5 & Single fixed run \\
        PushT & Source-CEM target contrast & TEST & 500 & 192 & 25/5 & 3 trajectory-field seeds \\
        PushT & Ranking attribution & TEST & 500 & 0/192 & 25/5 & Same 3 trajectory-field seeds \\
        \midrule
        TwoRoom & Action-space CEM & Strict TEST & 500 & 192 & 25/25 & Single fixed run \\
        TwoRoom & Flow executors & Strict TEST & 500 & 192 & 25/25 & 3 trajectory-field seeds \\
        TwoRoom & Ranking attribution & Strict TEST & 500 & 0/192 & 25/25 & Same 3 trajectory-field seeds \\
        \midrule
        Cube & Ranking attribution & TEST & 1,000 & 0/192 & 25/5 & Fixed model/field; 3 source seeds \\
        Cube & Target attribution & TEST & 1,000 & 192 & 25/5 & Fixed model/field; 3 source seeds \\
        \bottomrule
    \end{tabular}
\end{table}

All downstream task-trained components exclude official TEST episodes from gradient updates. PushT and TwoRoom proposers exclude their TEST lists from training and validation. CompassFlow excludes TEST episodes from training and calibration; retrieval banks do likewise. Cube readouts use 528 non-TEST training episodes, disjoint from the 1,000-episode T1/T3 panel and the full 2,500-episode TEST pool. Released LeWM pretraining lacks episode-level provenance. PushT inherits fixed LeWM action-normalization statistics that include its TEST episodes; these statistics remain shared across downstream studies (Appendix~\ref{sec:appendix-flow}).

\subsection{Checkpoints and evaluation randomness}
\label{app:protocol-randomness}

PushT proposer T1 reports three training seeds; T2/T3 use the checkpoint with median validation MSE across them. All TwoRoom proposer studies use the same TEST-excluded seed-0 checkpoint. PushT and TwoRoom flow-based results report mean and sample standard deviation across three trajectory-field training seeds; repeated evaluations of a panel are not independent episodes. Cube attribution fixes the model and field and varies only source-sampling seeds 0, 1, and 2.

All rows in Table~\ref{tab:appendix-target-protocols} use evaluation seed 0. Official TEST PushT search and ranking use source-noise base seed 0. TwoRoom flow executors use source-sampling seed 0 across field checkpoints and reuse the initial source coordinates across replans within each window. Primary PushT/TwoRoom ranking pairs use source-sampling seed 0; additional unranked source-seed repeats are reported only for TwoRoom. Appendix~\ref{sec:appendix-executor-attribution} specifies these comparisons and distinguishes Cube's matched IID suite from its headline Source-CEM run.

\subsection{Budgets, replanning, and stopping}
\label{app:protocol-schedules}

One predictive evaluation scores one candidate with frozen LeWM, after source decoding for CompassFlow. Prediction horizon $H$ and execution prefix $h$ are measured in raw actions. The reported horizon is $H=b\kappa=25$, with $b=5$ and $\kappa=5$; execution and stopping depend on the experiment below.

\paragraph{Target-interface experiments.}
The 9,000-evaluation target-interface studies use 25-action execution prefixes (Table~\ref{tab:appendix-target-protocols}). For a fixed five-episode PushT batch $\mathcal B$, let $M_{\mathcal B}=\max_{i\in\mathcal B}(g_i-s_i)$ for demonstration start/goal indices $s_i,g_i$. PushT T1, T2, and T3 use action caps of $M_{\mathcal B}+50$, 250, and $M_{\mathcal B}+50+L$, respectively, where $L$ is the number of replacement actions. TwoRoom proposer T1/T2/T3 use caps of 150/250/165. These studies latch success and continue to the cap with auto-reset.

\paragraph{PushT controlled search.}
The matched 192-budget and shared-initialization coordinate comparisons execute five-action prefixes, refresh proposer targets at each replan, and disable cross-plan warm starts (Appendix~\ref{app:revision-coordinate}). Replanning occurs when the buffer empties or an auto-reset flushes it. Arms share ordered episodes and action caps; success is latched while the batched loop continues to the cap. The budget sweep retains this execution loop (Appendix~\ref{app:revision-budget}).

\paragraph{TwoRoom search and ranking.}
Executor and ranking studies use 25-action execution/target-refresh intervals and a 150-action cap (Table~\ref{tab:appendix-search-protocols}). Execution stops at target entry or the cap. The budget sweep retains this TwoRoom protocol (Appendix~\ref{app:revision-budget}).

\paragraph{Cube full-episode execution.}
Table~\ref{tab:appendix-target-protocols} lists the separate LeCompass and Direct/Fixed-25 configurations. T1/T3 use a 225-action cap. Native reach does not terminate execution: LeCompass stops at the third opening action's post-action check or the cap, whereas Direct and Fixed-25 run to the cap.

\FloatBarrier
\subsection{Baselines and matched attribution arms}

\paragraph{Fixed-temporal oracle.}
For demonstration start and goal indices $s,g$, Fixed-$k$ uses observations at $s+k,s+2k,\ldots<g$ and appends the final goal. Segment budgets equal successive demonstration-index differences from $s$. Elapsed environment time alone selects the active waypoint, clamped to the final goal after the schedule ends; reaching a waypoint does not advance it. Fixed-25 therefore has the same target schedule as Direct when $g-s\leq25$.

This test-time demonstration access is unavailable to Direct and LeCompass. On Cube's 1,000-episode panel, Fixed-25 uses 3,966 intermediate-or-final observations from those episodes' expert frames, not the separate 1,500-episode reserve partition. Fixed-$k$ is undefined for cross-episode T2 under this same-episode construction.

\paragraph{Predictive-ranking attribution.}
Paired arms use matched source-noise construction within each field checkpoint and source seed (Appendix~\ref{app:protocol-randomness}). At a matched planning call, unranked execution selects the first decoded candidate without scoring; ranked IID selection scores all 192 candidates and selects the lowest-cost one. Both retain the learned prior conditioned on frozen JEPA history. Decoded trajectories may differ after closed-loop histories diverge. In the complete-interface Cube suite, both arms additionally retain current- and goal-token readouts and the execution program. The bare-planner suite omits these components (Appendix~\ref{app:cube-bare-ranking}).

\paragraph{Cube scoring-target attribution.}
Both arms retain ranked IID sampling with 192 candidates, the execution program, and the same current- and final-goal-token readouts. Only the scoring reference changes: final-goal latent versus proposed intermediate latent. The complete three-arm chain and its separation from the headline run are reported in Appendix~\ref{sec:appendix-executor-attribution}.

\subsection{Perturbations and duration sensitivity}

Table~\ref{tab:appendix-t3-protocols} defines the realized-execution perturbation used for each domain's T3 results.

\begin{table}[!htbp]
    \centering
    \small
    \caption{\textbf{T3 perturbation protocols.} Perturbations act on realized execution while the controller continues to observe and update.}
    \label{tab:appendix-t3-protocols}
    \begin{tabular}{@{}lp{0.72\columnwidth}@{}}
        \toprule
        Domain & Perturbation \\
        \midrule
        PushT & Beginning at $\lfloor0.4\,t_g\rfloor$, replace 15 emitted actions with actions sampled uniformly between each environment action-space bound; the inner policy continues to observe and update. \\
        TwoRoom & The same 15-action perturbation beginning at $\lfloor0.4\,t_g\rfloor$, evaluated with the strict collision-and-slide criterion. \\
        Cube & After the stable-carry trigger, apply the bounded forced-opening intervention defined in Appendix~\ref{app:cube-execution}. \\
        \bottomrule
    \end{tabular}
\end{table}
\FloatBarrier

For PushT/TwoRoom, $t_g=g-s$ is the demonstration goal offset from evaluation start. Replacement actions use a perturbation-policy random-number generator seeded with evaluation seed 0, independently of checkpoint randomness. Cube's trajectory-triggered intervention uses no random sampling (Appendix~\ref{app:cube-execution}); all 1,000 episodes remain in the denominator, including those that never meet the stable-carry trigger.

The 15-action perturbation protocol was fixed before the longer-duration sensitivity sweep. Table~\ref{tab:appendix-pusht-severity} varies duration on the same 500-episode TEST panel, retaining start fraction 0.4 and adding the replacement count to the nominal action cap.

\begin{table}[!htbp]
    \centering
    \small
    \caption{\textbf{PushT T3 perturbation-duration sensitivity.} Successes out of 500 TEST episodes, with rates in parentheses. The prespecified primary protocol replaces 15 actions; the remaining rows vary duration at the same start fraction of 0.4.}
    \label{tab:appendix-pusht-severity}
    \begin{tabular}{@{}rrrr@{}}
        \toprule
        Replaced actions & Direct & Fixed-25 & State-indexed \\
        \midrule
        15 & 21 (4.2\%) & 101 (20.2\%) & 145 (29.0\%) \\
        30 & 20 (4.0\%) & 50 (10.0\%) & 88 (17.6\%) \\
        45 & 20 (4.0\%) & 30 (6.0\%) & 61 (12.2\%) \\
        90 & 14 (2.8\%) & 26 (5.2\%) & 23 (4.6\%) \\
        \bottomrule
    \end{tabular}
\end{table}
\FloatBarrier

State-indexed execution retains a positive point difference over Fixed-25 through 45 replacement actions; the difference is $-0.6$ percentage points at 90 actions.

\subsection{Success criteria and statistical reporting}
\label{app:protocol-success}

\paragraph{PushT.}
Native joint object-and-tool success is latched over post-action frames: combined pusher-and-block position error is below 20 pixels and wrapped block-angle error below $\pi/9$.

\paragraph{TwoRoom.}
The strict protocol models the agent as a disk, resolves motion by swept-disk collision detection and obstacle sliding, and requires agent-center distance to the target below 16 pixels.

\paragraph{Cube.}
Native reach is \(\operatorname{ever}(d_{\mathrm{target}}\leq 0.04\,\mathrm{m})\), including frame zero; contact, pick, and carry are not prerequisites. Semantic completion additionally requires finished release and target containment in the post-action observation after the third opening command (Appendix~\ref{app:cube-execution}).

Fixed-seed Cube outcome intervals are episode-level Wilson intervals. Unless otherwise specified, paired binary-success $p$-values use two-sided exact McNemar tests, pairing episodes or windows within each checkpoint and sampling seed. Bootstrap confidence-interval procedures are specified with the corresponding results.

\section{Target Proposal and Trajectory Search Implementation}

\subsection{Proposer training construction}
\label{sec:appendix-proposer}

The frozen encoder constructs latent sequences at stride five, matching the action-block size. Each sample contains three normalized history latents, a normalized later-goal latent, and span $\kappa$. The target shifts all three history positions by $H=5\kappa$ raw actions; Eq.~\ref{eq:proposer-loss} averages their squared errors. The goal is sampled at or beyond the latest shifted position, with terminal-goal probability $0.5$; context and goal remain in one trajectory. Table~\ref{tab:appendix-proposer-config} gives the supported spans, architecture, and optimization.

\begin{table*}[!htbp]
    \centering
    \small
    \caption{\textbf{Proposer training configuration.} The architecture and optimization settings are shared across PushT, TwoRoom, and Cube.}
    \label{tab:appendix-proposer-config}
    \begin{tabular}{@{}ll@{}}
        \toprule
        Component & Configuration \\
        \midrule
        Input & 3 history latents, later-goal latent, and span index \\
        Output & 3 shifted latent predictions \\
        Sequence model & 6-layer transformer, 16 heads, 64-D head \\
        Latent dimension / MLP width & 192 / 2,048 \\
        Span embedding & 32-D; $\kappa\in\{2,\ldots,7\}$ \\
        Dropout & 0.1 \\
        Optimizer & AdamW, LR $3\times10^{-4}$, weight decay $10^{-3}$ \\
        Schedule & 2 warmup epochs, then cosine decay \\
        Batch / training duration & 1,024 / 30 epochs \\
        Validation & 10\% episode split; checkpointing by validation MSE \\
        \bottomrule
    \end{tabular}
\end{table*}

Only the latest output is inverse-normalized with training-split statistics for frozen-LeWM scoring. All reported proposer-based controllers use $\kappa=5$; target refresh follows the protocol-specific schedule in Appendix~\ref{app:protocol-schedules}, which can differ from replanning cadence.

\FloatBarrier
\subsection{Trajectory-prior training and action mapping}
\label{sec:appendix-flow}

Normalized action chunks occupy a padded 35-action tensor; deployment uses the 25-action horizon $\kappa=5$. With the interpolation from Eq.~\ref{eq:flow-loss}, let $v^\star=\bar{\mathbf a}-\epsilon$, and expand the validity mask $M_\kappa$ across action channels, with $\bar M_\kappa=1-M_\kappa$. The implemented objective includes a padded-coordinate penalty:
\[
    \mathcal L_{\mathrm{flow}}=
    \frac{\|M_\kappa\odot(v_\phi-v^\star)\|_2^2}{\|M_\kappa\|_1}
    +0.01\,
    \frac{\|\bar M_\kappa\odot v_\phi\|_2^2}
         {\max(\|\bar M_\kappa\|_1,1)}.
\]
Source noise and interpolation time are sampled independently per chunk. Padded source coordinates are zero, and the mask is re-applied during transport. Table~\ref{tab:appendix-flow-config} gives training and sampling settings.

\begin{table*}[!htbp]
    \centering
    \scriptsize
    \caption{\textbf{Trajectory-field training configuration.} Action dimension is 2 for PushT and TwoRoom and 5 for Cube.}
    \label{tab:appendix-flow-config}
    \begin{tabular}{@{}p{0.28\textwidth}p{0.67\textwidth}@{}}
        \toprule
        Component & Configuration \\
        \midrule
        Conditioning & 3 normalized 192-D history latents; 32-D span embedding; no target input \\
        Velocity network & 512-D condition MLP + 128-D sinusoidal-time MLP + 2-layer 512-D action MLP \\
        Capacity / deployed horizon & 35 / 25 raw actions; supported $\kappa\in\{2,\ldots,7\}$; deployed $\kappa=5$ \\
        Optimization & Adam, LR $3\times10^{-4}$, weight decay 0; batch 256; 8,000 steps \\
        Objective & Masked rectified-flow MSE on valid positions + $0.01$ auxiliary padded-coordinate penalty \\
        Sampling & 8 Euler steps \\
        Training repeats & PushT and TwoRoom executor studies: field seeds 0, 1, and 2; Cube: fixed seed-0 field \\
        Checkpoint & Final training step; no early stopping \\
        \bottomrule
    \end{tabular}
\end{table*}

After eight Euler steps, the valid normalized chunk is inverse-scaled and processed by the domain's raw-candidate rule (Table~\ref{tab:appendix-action-bridge}). The resulting raw trajectory is retained for execution; a copy standardized in LeWM's convention is scored. The selected raw prefix then enters the execution interface.

\begin{table*}[!htbp]
    \centering
    \small
    \caption{\textbf{Pre-scoring action bridge.} Invert the trajectory-field scaler, apply the raw-candidate rule, then standardize a copy for frozen-LeWM scoring. The raw candidate is retained as the planned trajectory and passed to the execution interface after selection.}
    \label{tab:appendix-action-bridge}
    \begin{tabular}{@{}llll@{}}
        \toprule
        Domain & Dim. & Raw candidate rule & LeWM action mapping \\
        \midrule
        PushT & 2 & No candidate clipping & $(a-\mu_{\mathrm{wm}})/s_{\mathrm{wm}}$ \\
        TwoRoom & 2 & Clip each channel to $[-1,1]$ & $(a-\mu_{\mathrm{wm}})/s_{\mathrm{wm}}$ \\
        Cube & 5 & Clip each channel to $[-1,1]$ & $(a-\mu_{\mathrm{wm}})/s_{\mathrm{wm}}$ \\
        \bottomrule
    \end{tabular}
\end{table*}

TwoRoom and Cube fit field action scalers on non-TEST training splits and store them with field artifacts. PushT shares the inherited LeWM action statistics between field training and scoring. At evaluation, the LeWM standardizer is reconstructed from the fixed raw-data artifact recorded in run provenance, not refitted to the evaluation panel.

\subsection{Source-space predictive trajectory search}
\label{app:source-search}

PushT and TwoRoom executor studies compare ranked IID sampling with Source-CEM. Ranking attribution in all domains and Cube target attribution use ranked IID; headline Cube T1/T3 uses Source-CEM.

Initial banks draw IID standard-normal coordinates on valid positions, with padded positions fixed at zero. PushT and Cube allocate banks per planning call; TwoRoom allocates per window and reuses source coordinates across replans. Reused coordinates are decoded under the new observed history, not reused as action trajectories.

\paragraph{Ranked IID sampling.}
Decode 192 source tensors under the current history, score each once with frozen LeWM, and send the lowest-cost raw prefix to the execution interface. Unranked execution selects the first decoded candidate without scoring. Neither arm adapts its source distribution.

\paragraph{Source-CEM (default configuration).}
The default 192-evaluation budget is divided into three 64-candidate rounds:
\begin{enumerate}
    \item Decode and score the initial 64 sources; fit a diagonal Gaussian to the eight lowest-cost sources in that round.
    \item Clamp the fitted mean to $[-1,1]$ and standard deviations to $[0.3,1]$.
    \item In each of two adaptive rounds, evaluate eight standard-normal prior anchors and 56 adaptive candidates (the mean and 55 samples), clamping evaluated source coordinates to $[-3,3]$. Refit and clamp using that round's eight elites.
    \item Return the lowest-cost decoded trajectory across all three rounds.
\end{enumerate}
The fitted distribution is not warm-started across calls. Prior anchors retain base-distribution coverage; adaptive proposals concentrate evaluations around low-cost sources. The budget sweep scales candidate and elite counts as specified in Appendix~\ref{app:revision-budget}. The separate coordinate comparison in Appendix~\ref{app:revision-coordinate} uses shared Flow initialization and disables anchors and trust bounds in both arms; it does not redefine this default configuration.

\section{Cube Execution Program}
\label{sec:appendix-cube}
\label{app:cube-execution}

Cube's readouts, state machines, channel ownership, and recovery are specified below. Evidence predicates, temporal thresholds, and recovery limits were determined from development rollouts and fixed before evaluation.

\subsection{Readouts and their execution roles}

Acquisition predicts world-frame block-minus-end-effector XY displacement from current patch tokens. Placement uses current and cached final-goal tokens to predict target-minus-block XY displacement $\widehat{\Delta xy}$ and block height. The controller derives lateral commitment and set-down decisions from these estimates, using height for vertical correction and descent/set-down eligibility. Execution memory gates release from sustained contact and lateral alignment, independently of the placement controller's set-down phase.

Final-goal tokens are encoded once per episode and are independent of the intermediate planning target. One placement estimate is shared by memory and the placement controller. Readouts are adapted sequentially before evaluation and then fixed (Appendix~\ref{app:readout-adaptation}).

\subsection{State machines and nominal progression}

Table~\ref{tab:cube-state-machines} lists two coupled state machines. The ten-state execution memory (hereafter phase memory) tracks contact persistence, grip/release eligibility, and recovery. It reads the observed scalar proprioceptive gripper-contact signal and $\widehat{\Delta xy}$, not simulator block/target coordinates or evaluator success; held contact means a signal above 0.5. The eight-state placement controller implements spatial ownership and set-down geometry, not additional task-level subgoals.

Nominal execution follows \textsc{Acquire}$\rightarrow$\textsc{Carry}$\rightarrow$\textsc{TargetHold}$\rightarrow$\textsc{Release}. \textsc{Acquire} groups \textsc{Seek}, \textsc{AcquirePending}, and \textsc{AcquireAssist}; \textsc{Release} groups the opening sequence and terminal \textsc{Released} state. Other memory states handle loss, recovery, and exhaustion.

\begin{table*}[!htbp]
    \centering
    \small
    \caption{\textbf{Coupled Cube state machines.} Execution memory maintains interaction evidence and recovery state; the placement controller determines spatial action ownership. Narrative task phases group execution-memory states.}
    \label{tab:cube-state-machines}
    \begin{tabular}{@{}p{0.14\textwidth}p{0.55\textwidth}p{0.22\textwidth}@{}}
        \toprule
        Machine & States & \raggedright Primary responsibility \tabularnewline
        \midrule
        \raggedright Execution memory & \raggedright \textsc{Seek}, \textsc{AcquirePending}, \textsc{AcquireAssist}, \textsc{Carry}, \textsc{Lost}, \textsc{SeekRecovery}, \textsc{TargetHold}, \textsc{Release}, \textsc{Released}, \textsc{RecoveryExhausted} & \raggedright Contact persistence, grip/release gating, bounded recovery \tabularnewline
        \addlinespace
        \raggedright Placement controller & \raggedright \textsc{PrePick}, \textsc{Transport}, \textsc{Center}, \textsc{Descend}, \textsc{CommitSetDown}, \textsc{AwaitHandover}, \textsc{Release}, \textsc{Stabilize} & \raggedright Spatial channel ownership and set-down realization \tabularnewline
        \bottomrule
    \end{tabular}
\end{table*}

\paragraph{Confirmation and release.}
Within one consecutive held-contact run, three observations activate bootstrap grip protection and twelve confirm \textsc{Carry}; earlier loss resets acquisition. Assisted acquisition expires after 60 steps and returns to \textsc{AcquirePending}. In \textsc{Carry}, held contact and $\|\widehat{\Delta xy}\|_2\leq0.012\,\mathrm{m}$ for two consecutive observations confirm \textsc{TargetHold}. Six further observations satisfying the same predicate initiate release; failure resets this dwell counter.

\textsc{Released} denotes a completed release-command sequence, not semantic success. T1 disables recovery, allows one carry activation, and stops after the third opening action or at timeout. Semantic completion is checked in that action's post-action observation, with no extra settling window (Appendix~\ref{app:protocol-success}).

\subsection{Action ownership and actuator order}

LeWM scores five-channel planner candidates before execution-side assignments. After trajectory selection, the placement controller assigns action channels according to Table~\ref{tab:appendix-cube-channel-ownership}. In \textsc{PrePick}, acquisition supplies $x,y$. The controller/actuator-composed action is not re-scored.

\begin{table*}[!htbp]
    \centering
    \small
    \caption{\textbf{Cube action-channel ownership before actuator overrides.} In \textsc{PrePick}, acquisition supplies $x,y$; the planner supplies $z$, yaw, and the base gripper action. In the other seven phases, placement supplies all but yaw. Grip-floor and release actuators act afterward.}
    \label{tab:appendix-cube-channel-ownership}
    \begin{tabular}{@{}lccccc@{}}
        \toprule
        Placement phase & $x$ & $y$ & $z$ & yaw & Gripper base \\
        \midrule
        \textsc{PrePick} & Acquisition & Acquisition & Planner & Planner & Planner \\
        Other seven phases & Placement & Placement & Placement & Planner & Placement \\
        \bottomrule
    \end{tabular}
\end{table*}

After channel assignment, the grip-floor actuator enforces a fixed lower bound on the gripper-closing command in \textsc{AcquireAssist}, \textsc{Carry}, and \textsc{TargetHold} to help maintain the grasp; otherwise the planner's pre-pick gripper value can pass through. The release actuator then overrides only the gripper with $-1.0$ for three actions. Release and grip-floor states are disjoint. The nominal order is planner, placement controller, grip floor, release actuator, then environment. Appendix~\ref{sec:appendix-cube-breakdowns} reports actual T1 value passthrough separately from assigned ownership.

\subsection{Bounded recovery}

Recovery-enabled T3 retains the learned components and nominal parameters, permits one recovery attempt, and allows at most two carry activations. Only \textsc{Carry} admits recovery: three consecutive unheld-contact observations lead through \textsc{Lost} to \textsc{SeekRecovery} if budget remains. Rollback atomically discards the unexecuted prefix and resets the placement controller to \textsc{PrePick} and the acquisition servo, preserving environment time, observed latent history, and outer-loop task progress. A new contact segment with twelve confirming observations activates the second \textsc{Carry} segment and resumes replanning. Loss after \textsc{TargetHold} does not enter recovery; exhaustion enters absorbing \textsc{RecoveryExhausted}.

\subsection{T3 forced-opening intervention}
\label{app:cube-perturbation}

After five consecutive \textsc{Carry} steps with held contact, the perturbation controller overrides only the gripper to $-1.0$, after nominal controllers and actuators and before environment execution. Each new observation reports contact from the preceding action. Before issuing another forced action, the override stops if three consecutive unheld observations confirm loss, the phase leaves \textsc{Carry}, or three forced-opening actions have been emitted. Reaching the cap without confirmed loss is a failed induction; later actions resume the nominal chain. The intervention uses no random sampling.

\section{Closed-Loop Readout Adaptation and Diagnostics}
\label{app:readout-adaptation}

This section supports the spatial-feedback analysis in Sec.~\ref{sec:analysis-execution-state}; the execution rules are specified in Appendix~\ref{sec:appendix-cube}.

\subsection{Adaptation and checkpoint selection}

Readout adaptation uses 528 training, 113 validation, and 113 held-out episodes. Expert supervision and rollout aggregation use training episodes; validation states support diagnostics and selection, not gradient updates. The held-out partition is unused here. The JEPA encoder and predictor remain frozen.

Acquisition starts with expert frames, then combines expert data with rollout states sampled uniformly across no-correction, pixel-based reference, and initial token-based controllers. The pixel reference estimates geometry directly from images. Further acquisition adaptation emphasizes near-field and false-zero hard negatives while retaining successful and far-field frames. Acquisition is then frozen while placement is adapted on expert frames and training rollouts from the initial token-based, pixel-based, and oracle controllers. Separate validation rollouts from these three sources support placement diagnostics and selection. Table~\ref{tab:readout-implementation} specifies the deployed heads.

\begin{center}
\begin{minipage}{\textwidth}
    \small
    \refstepcounter{table}\label{tab:readout-implementation}
    \noindent Table~\thetable: \textbf{Deployed Cube readouts.} Simulator state supplies training labels; deployed readouts consume frozen visual tokens. Both heads are fixed before TEST evaluation.

    \vspace{3pt}
    \centering
    \begin{tabular}{@{}p{0.24\textwidth}p{0.31\textwidth}p{0.31\textwidth}@{}}
        \toprule
        & Acquisition (R2) & Placement (P1) \\
        \midrule
        \raggedright Input and target & Current $16\!\times\!16\!\times\!192$ patch tokens; world-frame $xy_{\mathrm{block}}-xy_{\mathrm{effector}}$ (2 values). & Current and cached final-goal token grids; world-frame $xy_{\mathrm{target}}-xy_{\mathrm{block}}$ and block height above rest (3 values). \\
        Network & $1\!\times\!1$ convolutions $192\!\to\!128\!\to\!32$, spatial softmax to 32 keypoint coordinates (64 values), then a 256--256 MLP. & The same spatial trunk for each input, with separate weights; concatenated features feed a 256--256 MLP. No learned phase-gate head. \\
        \raggedright Loss and optimization & Huber loss ($\delta=1$) on targets divided by $0.05\,\mathrm m$; AdamW, learning rate $10^{-3}$, weight decay $10^{-4}$, OneCycle schedule; 8,000 steps, batch 128. & Same scaled Huber loss and AdamW settings, 8,000 steps, batch 128; constant learning rate. \\
        \raggedright Training-state sampling & Choose among four equally weighted sources (expert, no-correction, pixel-controller, token-controller), then sample an episode and a frame; within-episode frame weights favor near-field and false-zero cases. & Choose among four equally weighted sources (expert, initial-token, pixel-controller, oracle), then sample an episode, phase, and frame. \\
        \bottomrule
    \end{tabular}
\end{minipage}
\end{center}

The deployed acquisition and placement checkpoints were selected using offline validation diagnostics. Selection was finalized before the corresponding closed-loop outcomes were observed, and the selected readouts remained fixed throughout formal evaluation.

\subsection{Readout errors across evaluation distributions}

Expert-frame diagnostics use the 113-episode validation partition: 3,830 acquisition and 9,469 placement frames. Controller-induced diagnostics use 5,338 frames from 40 validation episodes under the expert-frame-trained acquisition readout, and 5,580 failure-state frames from 31 validation episodes under the initial placement readout. These stored states support offline diagnostics and selection, not training. Table~\ref{tab:readout-adaptation-diagnostics} compares readouts on identical frames within each rollout-state pair.

\begin{table*}[!htbp]
    \centering
    \small
    \caption{\textbf{Readout errors on expert and controller-induced states.} Expert and rollout rows use distinct distributions; readouts within each rollout pair use identical frames. p50/p90 summarize per-frame world-frame XY Euclidean errors (mm): block minus effector for acquisition, target minus block for placement.}
    \label{tab:readout-adaptation-diagnostics}
    \begin{tabular}{@{}lllrrr@{}}
        \toprule
        Stage & Readout & Evaluation states & Episodes & p50 & p90 \\
        \midrule
        Acquisition & Expert-frame trained & Expert validation & 113 & 2.62 & 5.22 \\
        Acquisition & Rollout adapted & Expert validation & 113 & 3.81 & 10.12 \\
        Acquisition & Expert-frame trained & Self-induced rollout & 40 & 62.98 & 133.09 \\
        Acquisition & Rollout adapted & Same 5,338 rollout frames & 40 & 9.23 & 42.31 \\
        \midrule
        Placement & Initial & Expert validation & 113 & 5.71 & 11.17 \\
        Placement & Rollout adapted & Expert validation & 113 & 5.90 & 11.22 \\
        Placement & Initial & Self-induced failure states & 31 & 190.46 & 733.50 \\
        Placement & Rollout adapted & Same 5,580 rollout frames & 31 & 11.86 & 21.30 \\
        \bottomrule
    \end{tabular}
\end{table*}

Across no-correction, pixel-reference, and initial-token rollout states, acquisition median error changes from 49.44, 14.09, and 62.98 mm to 10.10, 7.34, and 9.23 mm, respectively.

On the same 5,338 acquisition rollout frames, wrong-direction predictions fall from 2,420 (45.3\%) to 118 (2.2\%). Wrong direction means a negative dot product with the true correction. Predicted or true vectors with norm at most $10^{-9}\,\mathrm{m}$ are excluded; all 5,338 frames remain eligible for both readouts.

\subsection{Closed-loop validation}

On the same 113-episode validation panel, acquisition adaptation increases confirmed picks from 47/113 (41.6\%; 95\% Wilson interval $[32.9,50.8]$) to 93/113 (82.3\%; $[74.2,88.2]$), with 48 improved and two worsened pairs ($p=2.27\times10^{-12}$).

Holding adapted acquisition fixed yields the same 93 confirmed-pick episodes for both placement configurations. Within this cohort, native reach increases from 11/93 (11.8\%; $[6.7,19.9]$) to 92/93 (98.9\%; $[94.2,99.8]$), with 81 improved and zero worsened pairs ($p=8.27\times10^{-25}$). Off-target set-downs decrease from 36/93 (38.7\%; $[29.4,48.9]$) to 3/93 (3.2\%; $[1.1,9.1]$). Off-target set-down is one of six mutually exclusive placement-outcome categories; the six categories sum to 93 episodes in each configuration. Native reach and off-target set-down are not mutually exclusive.

\subsection{Representation and input interventions}
\label{app:readout-representation}

\begin{figure}[!htbp]
    \centering
    \includegraphics[width=0.83\columnwidth]{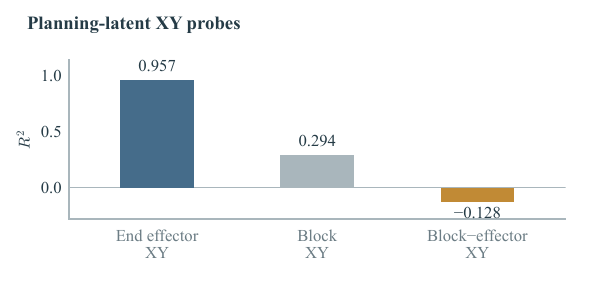}
    \caption{\textbf{Projected-CLS planning-latent XY probes.} MLP $R^2$ on the same 14,656 validation frames as Table~\ref{tab:appendix-spatial-probes}. The probes assess XY readout, not height or action-conditioned ranking; poor decoding does not establish that the latent lacks the information.}
    \label{fig:appendix-planning-latent-probes}
\end{figure}

\paragraph{Planning-latent probes.}
Ridge and MLP probes use frozen 192-dimensional projected CLS latents, with 71,232 training and 14,656 validation frames. Both decode end-effector XY more accurately than block-relative XY (Table~\ref{tab:appendix-spatial-probes} and Fig.~\ref{fig:appendix-planning-latent-probes}); height is not tested.

\begin{table}[!htbp]
    \centering
    \small
    \caption{\textbf{Spatial probes of the frozen planning latent.} All scores use the same 14,656 validation frames. Median errors are world-frame XY Euclidean distances.}
    \label{tab:appendix-spatial-probes}
    \begin{tabular}{@{}lrrr@{}}
        \toprule
        Prediction target & Ridge $R^2$ & MLP $R^2$ & MLP median error (mm) \\
        \midrule
        End-effector XY & 0.946 & 0.957 & 16.3 \\
        Block XY & 0.334 & 0.294 & 89.7 \\
        Block minus end-effector XY & $-0.036$ & $-0.128$ & 90.7 \\
        \bottomrule
    \end{tabular}
\end{table}

For relative XY, a zero-displacement predictor has median error 73.1 mm, versus 5.68 mm for a development-only pixel CNN on the same frames. These are median-distance comparisons; $R^2$ instead compares mean-squared error with a mean-label predictor.

\paragraph{Patch-token localization.}
Each intervention swaps one $16\times16$ pixel cell between two frames, re-encodes both, and ranks 256 tokens by change magnitude. Five of twelve swaps have identical pixels and are excluded. The corresponding token ranks first in all seven non-null swaps.

\paragraph{Readout input interventions.}
Initial, pre-adaptation readouts are tested with changed inputs but fixed labels and weights. Independently shuffling token positions within each sample raises acquisition expert-frame p90 error from 5.22 to 253.32 mm. For placement, swapping current/goal inputs, replacing the goal with the current input, and shuffling tokens raise p90 error from 11.17 mm to 149.33, 126.19, and 188.02 mm, respectively.

\section{Frozen-Latent Target-Geometry Diagnostics}
\label{sec:appendix-resolvability}

\subsection{Distance distributions and summary scales}

The diagnostic in Sec.~\ref{sec:analysis-resolvability} uses three unsquared $L_2$ distance distributions in each frozen checkpoint. $\mathcal A$ measures within-episode anchor-to-successor distances over one latent step (five raw frames). $\mathcal G$ uses uniformly sampled pairs of distinct retained anchor-frame entries, including within- and cross-episode pairs. $\mathcal E$ measures error at the proposer's latest inverse-normalized output:
\begin{equation}
    d_t^{\mathrm{prop}}=\left\|P_\psi(\mathbf z_t^{\mathrm{hist}},z_{\mathrm{final}},\kappa)-z_{t+\kappa}\right\|_2,
    \qquad \kappa=5.
\end{equation}
Here $t$ and $\kappa$ index latent steps. History contains the anchor and two preceding latents; the supervised future is 25 raw frames ahead, whereas $\mathcal A$ spans five.

Define $A=\operatorname{median}(\mathcal A)$, $G=q_{0.95}(\mathcal G)$, and $E=\operatorname{median}(\mathcal E)$. Normalizing all distances by $G$ preserves within-checkpoint ordering and gives the summary ratios
\begin{equation}
    R_{\mathrm{usable}}=1-\frac{A}{G},
    \qquad
    R_{\mathrm{error}}=\frac{E/G}{1-A/G}.
    \label{eq:appendix-resolvability-conversion}
\end{equation}
Thus $R_{\mathrm{error}}=E/(G-A)$ is the main-text distance ratio, not a dynamical-reachability test.

\subsection{Anchor sampling and uncertainty}

For each of 754 validation episodes per environment, we request eight evenly spaced anchors with the required history and five-step future. Shorter valid ranges contribute every available position without duplication. Each anchor contributes one $\mathcal A$ and one $\mathcal E$ sample; equally many global pairs form $\mathcal G$. Each distribution has 5,994 samples for PushT, 6,032 for Cube, and 5,845 for TwoRoom.

Uncertainty uses 2,000 episode-level bootstrap resamples. Each resample retains the selected episodes' anchors, redraws global pairs from the resampled anchor pool, and recomputes both ratios from its medians and global-pair 95th percentile. The reported 95\% intervals are the 2.5th and 97.5th percentiles of these ratio draws.

\subsection{Distributions and cross-domain comparison}

\begin{figure*}[!htbp]
    \centering
    \includegraphics[width=0.82\textwidth]{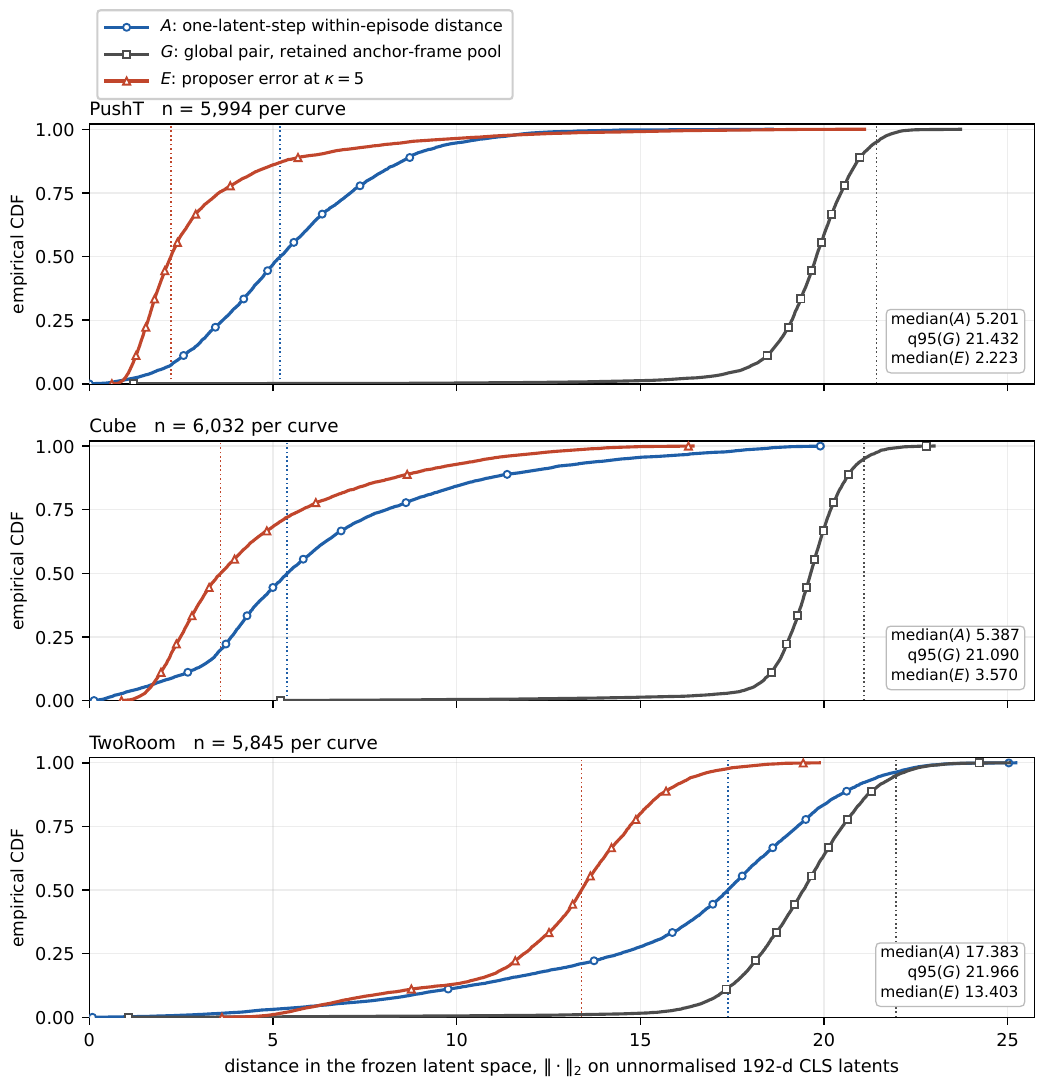}
    \caption{\textbf{Frozen-latent distance distributions.} Empirical CDFs of one-latent-step distance $\mathcal A$, global-pair distance $\mathcal G$, and proposer error $\mathcal E$ (labeled A, G, and E) for independently trained domain checkpoints. Distances are unsquared $L_2$ in unnormalized 192-D projected CLS space. Dotted lines mark $A=\operatorname{median}(\mathcal A)$, $G=q_{0.95}(\mathcal G)$, and $E=\operatorname{median}(\mathcal E)$.}
    \label{fig:latent-distance-raw}
\end{figure*}

\begin{figure*}[t]
    \centering
    \includegraphics[width=0.82\textwidth]{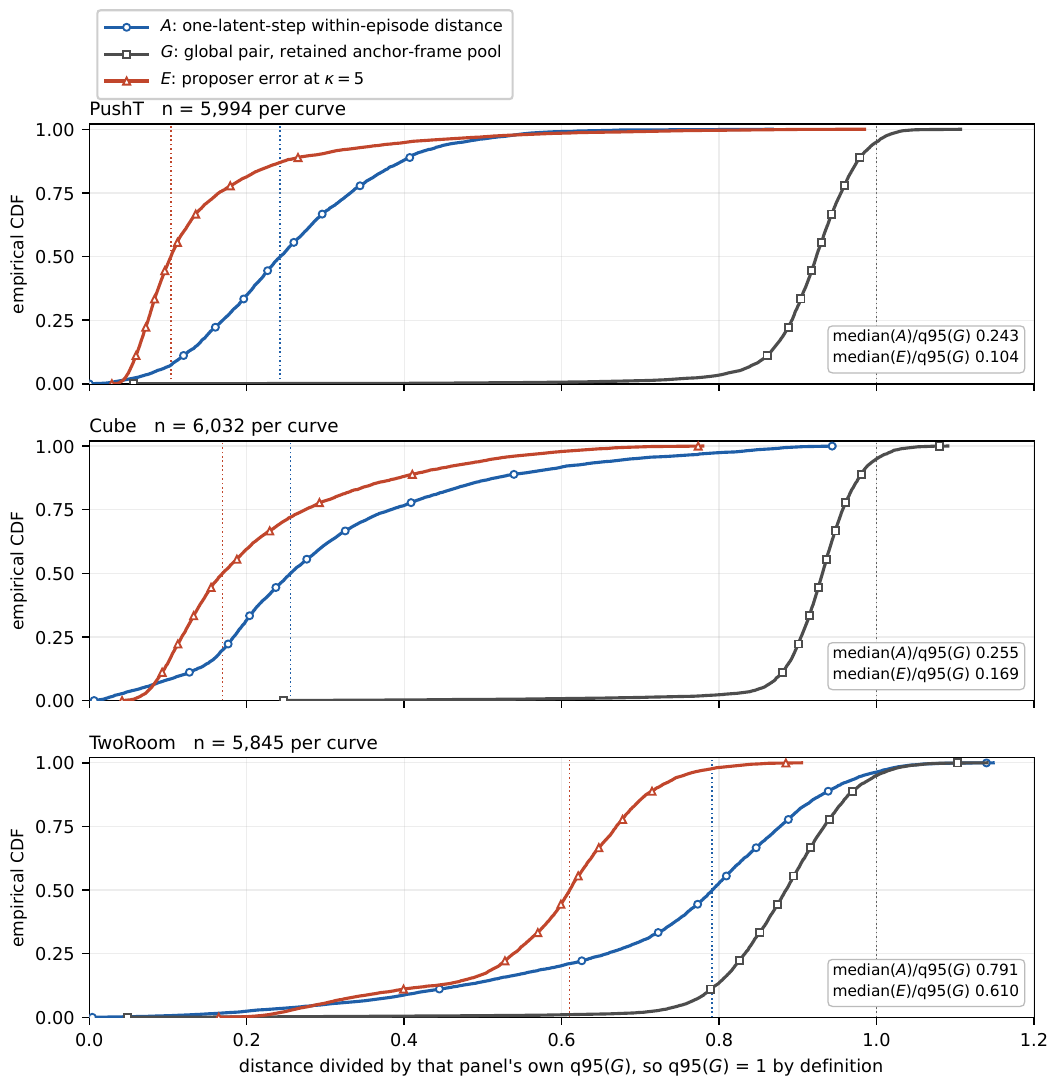}
    \caption{\textbf{Checkpoint-normalized distances.} All $\mathcal A$, $\mathcal G$, and $\mathcal E$ distances are divided by the checkpoint-specific $q_{0.95}(\mathcal G)$. This sets the 95th-percentile global-pair reference to one for cross-checkpoint scale comparison without changing within-checkpoint distance ordering.}
    \label{fig:latent-distance-normalized}
\end{figure*}

Figures~\ref{fig:latent-distance-raw} and~\ref{fig:latent-distance-normalized} show raw and checkpoint-normalized empirical CDFs in the unnormalized 192-dimensional projected CLS space. Table~\ref{tab:appendix-target-geometry} reports the summary ratios. PushT and Cube have median local-transition distances and proposer errors well below the global-pair scale; TwoRoom has narrower local-to-global separation relative to proposer error.

\begin{table}[!htbp]
    \centering
    \small
    \caption{\textbf{Relative target-geometry scales.} Ratios use unrounded statistics; brackets are 95\% episode-bootstrap intervals. $R_{\mathrm{error}}=E/(G-A)$ is a distance ratio and may exceed one.}
    \label{tab:appendix-target-geometry}
    \begin{tabular}{@{}lrrll@{}}
        \toprule
        Environment & $A/G$ & $E/G$ & $R_{\mathrm{usable}}$ [95\% CI] & $R_{\mathrm{error}}$ [95\% CI] \\
        \midrule
        PushT & 0.243 & 0.104 & 0.7573 [0.7515, 0.7620] & 0.1370 [0.1327, 0.1422] \\
        Cube & 0.255 & 0.169 & 0.7446 [0.7404, 0.7470] & 0.2274 [0.2231, 0.2329] \\
        TwoRoom & 0.791 & 0.610 & 0.2087 [0.2026, 0.2152] & 2.9238 [2.8229, 3.0240] \\
        \bottomrule
    \end{tabular}
\end{table}

\FloatBarrier
\section{Executor and Attribution Breakdowns}
\label{sec:appendix-executor-attribution}

The comparisons below separate search configuration, scoring target, and predictive ranking (Secs.~\ref{sec:results-executor} and~\ref{sec:analysis-attribution}). Protocols and source-bank construction are specified in Appendices~\ref{sec:appendix-protocol} and~\ref{app:source-search}. Repeated checkpoints or source seeds are not additional independent episodes.

\subsection{Official TEST PushT search and refinement}

Table~\ref{tab:appendix-pusht-test-search} uses the same ordered 500 TEST episodes, frozen LeWM and proposer, 192 evaluations per plan, and nominal five-action prefix. Both Source-CEM arms refresh targets every five actions and refit at each call; only the scoring target changes. Matched Action-CEM results are reported in Appendix~\ref{app:revision-coordinate}.

\begin{table*}[!htbp]
    \centering
    \small
    \setlength{\tabcolsep}{4pt}
    \caption{\textbf{Official TEST PushT scoring-target contrast.} Counts use the same 500 episodes within each field checkpoint. Differences are proposed minus final-goal Source-CEM in percentage points; 95\% intervals use 10,000 episode-paired bootstrap resamples.}
    \label{tab:appendix-pusht-test-search}
    \begin{tabular}{@{}crrl@{}}
        \toprule
        Field seed & Source proposed & Source final & Difference [95\% CI] \\
        \midrule
        0 & 264 & 76 & $+37.6$ $[32.6,42.6]$ \\
        1 & 274 & 64 & $+42.0$ $[36.8,47.2]$ \\
        2 & 249 & 68 & $+36.2$ $[31.0,41.4]$ \\
        \bottomrule
    \end{tabular}
\end{table*}

Ranked IID and proposed-target Source-CEM share the panel, checkpoint, model, proposer, scoring target, prefix, and per-plan budget; only source sampling is refined (Table~\ref{tab:appendix-pusht-test-refinement}). Across the three field checkpoints, success is $44.9\pm1.1$\% for ranked IID and $52.5\pm2.5$\% for Source-CEM (mean $\pm$ sample standard deviation).

\begin{table*}[!htbp]
    \centering
    \scriptsize
    \setlength{\tabcolsep}{3pt}
    \caption{\textbf{Official TEST PushT source refinement.} Each row pairs 500 episodes within a field checkpoint. Differences are Source-CEM minus ranked IID (percentage points); 95\% intervals use 10,000 episode-paired bootstrap resamples. Checkpoints are not pooled as independent episodes.}
    \label{tab:appendix-pusht-test-refinement}
    \begin{tabular}{@{}crrclc@{}}
        \toprule
        Field seed & Source-CEM & Ranked IID & Difference [95\% CI] & Source-only / IID-only & Exact $p$ \\
        \midrule
        0 & 264/500 (52.8\%) & 228/500 (45.6\%) & $+7.2$ $[2.0,12.4]$ & 106 / 70 & $.00815$ \\
        1 & 274/500 (54.8\%) & 227/500 (45.4\%) & $+9.4$ $[4.2,14.6]$ & 119 / 72 & $.000828$ \\
        2 & 249/500 (49.8\%) & 218/500 (43.6\%) & $+6.2$ $[0.6,11.4]$ & 107 / 76 & $.0263$ \\
        \bottomrule
    \end{tabular}
\end{table*}

\subsection{Official TEST TwoRoom executor and refinement}

On 500 strict TEST TwoRoom windows with the proposed target and a nominal 25-action prefix, ranked IID reaches $77.3\pm0.4$\% and Source-CEM $77.3\pm1.0$\% across three field checkpoints. Table~\ref{tab:appendix-tworoom-executor-seeds} gives per-checkpoint counts, including final-goal Source-CEM. Matched Source-CEM-minus-ranked-IID effects are $-0.8$, $0.0$, and $+0.8$ percentage points across field checkpoints 0--2 (Table~\ref{tab:appendix-tworoom-refinement}). A two-level sensitivity bootstrap with 200,000 resamples over field checkpoints and paired windows gives a 95\% interval of $[-2.3,2.3]$ points for the mean difference. Equal aggregate successes do not imply identical successful windows.

\begin{table*}[!htbp]
    \centering
    \scriptsize
    \setlength{\tabcolsep}{2pt}
    \caption{\textbf{Official TEST TwoRoom executor outcomes.} Counts use the same 500 strict windows for field seeds 0--2. Aggregates sum repeated evaluations, not independent windows. Source-sampling seed is 0.}
    \label{tab:appendix-tworoom-executor-seeds}
    \begin{tabular}{@{}llrrrr@{}}
        \toprule
        Domain ($N$ per seed) & Configuration & Seed 0 & Seed 1 & Seed 2 & Aggregate \\
        \midrule
        TwoRoom (500) & Ranked flow sampling + proposed latent & 385 & 389 & 386 & 1,160/1,500 \\
        TwoRoom (500) & Source-CEM + proposed latent & 381 & 389 & 390 & 1,160/1,500 \\
        TwoRoom (500) & Source-CEM + final-goal latent & 310 & 300 & 320 & 930/1,500 \\
        \bottomrule
    \end{tabular}
\end{table*}

\begin{table}[!htbp]
    \centering
    \small
    \setlength{\tabcolsep}{4pt}
    \caption{\textbf{Official TEST TwoRoom source refinement.} Source-CEM minus ranked IID on the same 500 strict TEST windows within each field checkpoint. Improved/worsened windows succeed only under Source-CEM/ranked IID. Intervals use 100,000 paired percentile-bootstrap resamples.}
    \label{tab:appendix-tworoom-refinement}
    \begin{tabular}{@{}lrrlr@{}}
        \toprule
        Field seed & Difference (pp) & Improved/worsened & 95\% CI (pp) & $p$ \\
        \midrule
        0 & $-0.8$ & 44/48 & $[-4.6,3.0]$ & .7547 \\
        1 & $0.0$ & 45/45 & $[-3.8,3.8]$ & 1.0000 \\
        2 & $+0.8$ & 44/40 & $[-2.8,4.4]$ & .7436 \\
        \bottomrule
    \end{tabular}
\end{table}

\subsection{Predictive-ranking attribution on PushT and TwoRoom}

\paragraph{PushT.}
Table~\ref{tab:appendix-ranking-pusht-test} compares proposed-target IID best-of-192 selection with matched unranked execution on the same 500 TEST episodes, within each field checkpoint. Both retain the encoder and prior and share source coordinates at matched planning points; decoded trajectories can diverge with observed histories. Only the ranked arm scores candidates.

\begin{table*}[!htbp]
    \centering
    \scriptsize
    \setlength{\tabcolsep}{3pt}
    \caption{\textbf{Official TEST PushT predictive ranking.} Rows pair 500 episodes within each field checkpoint at source seed 0. Differences are ranked minus unranked (percentage points). Intervals use 10,000 episode-paired bootstrap resamples. Repeats are not independent episodes.}
    \label{tab:appendix-ranking-pusht-test}
    \begin{tabular}{@{}crrclc@{}}
        \toprule
        Field seed & Best of 192 & Unranked & Difference [95\% CI] & Ranked-only / unranked-only & Exact $p$ \\
        \midrule
        0 & 228/500 (45.6\%) & 27/500 (5.4\%) & $+40.2$ $[35.6,44.8]$ & 212 / 11 & $2.07\times10^{-49}$ \\
        1 & 227/500 (45.4\%) & 27/500 (5.4\%) & $+40.0$ $[35.2,44.6]$ & 211 / 11 & $3.93\times10^{-49}$ \\
        2 & 218/500 (43.6\%) & 16/500 (3.2\%) & $+40.4$ $[36.0,44.8]$ & 206 / 4 & $9.76\times10^{-56}$ \\
        \bottomrule
    \end{tabular}
\end{table*}

\paragraph{TwoRoom.}
Table~\ref{tab:appendix-ranking-tworoom} pairs proposed-target IID ranking and unranked execution at cadence 25 and source seed 0 on 500 strict TEST windows. These attribution runs are separate from the executor runs in Table~\ref{tab:appendix-tworoom-executor-seeds}.

\begin{table*}[!htbp]
    \centering
    \scriptsize
    \setlength{\tabcolsep}{2pt}
    \caption{\textbf{TwoRoom predictive ranking at cadence 25.} Matched source construction and seed 0 on 500 strict TEST windows; ranked IID selection uses the proposed target. Bootstrap 95\% intervals pair windows within each field checkpoint.}
    \label{tab:appendix-ranking-tworoom}
    \begin{tabular}{@{}lrrrrr@{}}
        \toprule
        Field seed & Best of 192 & No predictive ranking & Paired difference & Paired bootstrap 95\% CI & $p$ \\
        \midrule
        0 & 387/500 (77.4\%) & 152/500 (30.4\%) & +47.0 pp & $[42.6,51.4]$ & $6.49\times10^{-68}$ \\
        1 & 390/500 (78.0\%) & 178/500 (35.6\%) & +42.4 pp & $[37.8,47.0]$ & $1.30\times10^{-57}$ \\
        2 & 388/500 (77.6\%) & 177/500 (35.4\%) & +42.2 pp & $[37.6,46.8]$ & $1.49\times10^{-54}$ \\
        \midrule
        Mean$\pm$SD & 77.7$\pm$0.3\% & 33.8$\pm$2.9\% & +43.9$\pm$2.7 pp & -- & -- \\
        \bottomrule
    \end{tabular}
\end{table*}

Unranked source seeds 0--2 yield success ranges of 11.4--30.4\%, 20.8--35.6\%, and 16.4--35.4\% for field checkpoints 0--2. The prespecified matched comparison uses source seed 0.

\subsection{Cube predictive-ranking and target attribution with the complete execution interface}

The Cube IID suite has three arms: first-candidate unranked execution ($A$), final-goal best-of-192 ranking ($B$), and proposed-target best-of-192 ranking ($C$). All share the 1,000-episode panel, one fixed model and field, and the execution interface. Thus $B-A$ adds final-goal ranking, $C-B$ changes its scoring target, and $C-A$ measures the net effect relative to unranked execution.

Source seeds 0--2 vary sampling only; seed 0 is primary and seeds 1--2 test directional sensitivity. The headline T1 run instead uses Source-CEM with source-seed base 7.

\begin{table*}[!htbp]
    \centering
    \scriptsize
    \setlength{\tabcolsep}{2pt}
    \caption{\textbf{Cube IID ranking and scoring-target attribution with the complete execution interface.} Native 4-cm reach on the same 1,000 episodes. Intervals are episode-paired within source seed. Mean$\pm$SD summarizes these three source-seed runs.}
    \label{tab:appendix-ranking-cube}
    \textbf{Absolute native-reach rates}\par\smallskip
    \begin{tabular}{@{}lrrr@{}}
        \toprule
        Source-sampling seed & A: Unranked & B: Final goal & C: Proposed target \\
        \midrule
        0 & 853/1000 (85.3\%) & 846/1000 (84.6\%) & 872/1000 (87.2\%) \\
        1 & 865/1000 (86.5\%) & 837/1000 (83.7\%) & 872/1000 (87.2\%) \\
        2 & 873/1000 (87.3\%) & 852/1000 (85.2\%) & 874/1000 (87.4\%) \\
        \midrule
        Mean$\pm$SD & 86.4$\pm$1.0\% & 84.5$\pm$0.8\% & 87.3$\pm$0.1\% \\
        \bottomrule
    \end{tabular}
    \par\medskip
    \textbf{Episode-paired effects: difference [95\% CI], $p$}\par\smallskip
    \begin{tabular}{@{}llll@{}}
        \toprule
        Source-sampling seed & B$-$A: final-goal ranking & C$-$B: target replacement & C$-$A: net vs. unranked \\
        \midrule
        0 & $-0.7$ $[-2.8,1.3]$, $p=.566$ & $+2.6$ $[0.7,4.5]$, $p=.0095$ & $+1.9$ $[0.0,3.8]$, $p=.059$ \\
        1 & $-2.8$ $[-4.8,-0.9]$, $p=.0078$ & $+3.5$ $[1.4,5.6]$, $p=.0016$ & $+0.7$ $[-1.1,2.5]$, $p=.52$ \\
        2 & $-2.1$ $[-3.8,-0.4]$, $p=.020$ & $+2.2$ $[0.2,4.1]$, $p=.033$ & $+0.1$ $[-1.9,2.0]$, $p=1.0$ \\
        \midrule
        Mean$\pm$SD & $-1.9\pm1.1$ pp & $+2.8\pm0.7$ pp & $+0.9\pm0.9$ pp \\
        \bottomrule
    \end{tabular}
\end{table*}

Final-goal ranking has negative native-reach point differences, while target replacement has positive differences, in all three runs (Table~\ref{tab:appendix-ranking-cube}). Semantic-completion differences follow the same directions: $B-A$ is $-1.9,-3.0,-1.8$ points; $C-B$ is $+3.0,+3.1,+2.0$; and $C-A$ is $+1.1,+0.1,+0.2$. Small net effects coexist with high success under the shared prior and goal-aware execution interface. Predictive ranking without the execution interface is evaluated separately in Appendix~\ref{app:cube-bare-ranking}.

\FloatBarrier
\section{Cube Outcome and Recovery Breakdowns}
\label{sec:appendix-cube-breakdowns}

\subsection{T1 outcomes and trajectory diagnostics}

Both fixed 500-episode partitions use the same frozen full-episode T1 configuration. Table~\ref{tab:appendix-cube-partitions} gives partition outcomes and their count-based micro-average; native reach is 86.9\% with a 95\% episode-level Wilson interval of $[84.7,88.9]$\%.

\begin{table}[!htbp]
    \centering
    \small
    \caption{\textbf{Cube T1 partition outcomes.} Native reach and semantic completion are reported separately and combined by count across the two fixed 500-episode partitions.}
    \label{tab:appendix-cube-partitions}
    \begin{tabular}{@{}lrr@{}}
        \toprule
        Partition & Native reach & Semantic completion \\
        \midrule
        Official main & 440/500 (88.0\%) & 433/500 (86.6\%) \\
        Legacy windowed-protocol partition & 429/500 (85.8\%) & 427/500 (85.4\%) \\
        \midrule
        Micro-average & 869/1,000 (86.9\%) & 860/1,000 (86.0\%) \\
        \bottomrule
    \end{tabular}
\end{table}

Table~\ref{tab:appendix-cube-common-funnel} separates contact, geometric pick, lift, and reach. Lift is relative to the demonstration's fifth-percentile block height; maximum lift uses the retrospective descriptive threshold $>2$ cm. Rollout-earned reach requires starting outside the target and first entering after execution begins, with all 1,000 episodes as denominator.

\begin{table*}[!htbp]
    \centering
    \scriptsize
    \setlength{\tabcolsep}{2pt}
    \caption{\textbf{Cube T1 trajectory diagnostics on the common 1,000-episode panel.} Native reach includes frame zero; rollout-earned reach first occurs after execution begins. Geometric true pick requires contact $>0.5$ and block lift $\geq5$ cm in the same observation.}
    \label{tab:appendix-cube-common-funnel}
    \begin{tabular}{@{}lrrrrr@{}}
        \toprule
        Method & Any contact & Geometric true pick & Maximum lift $>2$ cm & Native reach & Rollout-earned reach \\
        \midrule
        Direct & 569 & 0 & 0 & 37 & 1 \\
        Fixed-25 & 643 & 1 & 1 & 43 & 7 \\
        LeCompass & 959 & 820 & 893 & 869 & 833 \\
        \bottomrule
    \end{tabular}
\end{table*}

Baseline failures occur primarily between contact and geometric pick. Paired native-reach tests give $p<10^{-200}$ for LeCompass versus either baseline; Fixed-25 versus Direct is unresolved ($p=.0703$). Rollout-earned reach is 83.3\%, 0.1\%, and 0.7\% for LeCompass, Direct, and Fixed-25, respectively, excluding frame-zero entries as an explanation of the gap. The baseline-to-LeCompass nominal $B/h$ ratio is $9.4\times$ (Table~\ref{tab:cube-results}(a)).

The 1,000 T1 episodes comprise 860 semantic completions, 80 with contact but no geometric pick, 41 with no contact, 17 with post-pick drop or absent carry, and two with native reach but incomplete release. Completion is assigned first; other episodes follow the first failed check in the order contact, geometric pick, retained carry, native reach, release, and containment. These mutually exclusive classes cover the full panel, not native-reach successes. Geometric pick is not part of the completion criterion.

Table~\ref{tab:appendix-cube-motion} summarizes block motion on the same panel.

\begin{table}[!htbp]
    \centering
    \small
    \caption{\textbf{Cube T1 block motion.} Medians over all 1,000 episodes, in meters. Tiny nonzero baseline values reflect the simulator numerical floor, not earned lift.}
    \label{tab:appendix-cube-motion}
    \begin{tabular}{@{}lrr@{}}
        \toprule
        Method & Block XY path p50 & Maximum block lift p50 \\
        \midrule
        Direct & $5.7\times10^{-14}$ & $4.0\times10^{-5}$ \\
        Fixed-25 & $8.7\times10^{-14}$ & $4.0\times10^{-5}$ \\
        LeCompass & 0.286195 & 0.103273 \\
        \bottomrule
    \end{tabular}
\end{table}

\subsection{Realized action passthrough}
Of 82,910 headline T1 actions, 57,520 (69.4\%) occur in the placement controller's \textsc{PrePick} state; other placement-controller states account for 30.6\%. Planner $z$ passes through on all pre-pick actions and yaw on all 82,910 actions. Pre-pick gripper commands pass through on 51,531/57,520 actions; the grip floor changes 5,989. Planner $x,y$ values almost never pass through because readouts supply them. These are observed value-passthrough counts, not causal channel contributions. The bare-planner ranking comparison passes all action channels through unchanged (Appendix~\ref{app:cube-bare-ranking}).

Among 820 geometric-pick episodes, median first-pick time is step 34 and the median fraction of observations at or after pick is 41.7\%, including the pick observation.

\subsection{Height-informed simple-control reference}
\label{sec:appendix-cube-simple-control}

To test whether a simple controller can operate the Cube execution stack without learned action proposals or predictive scoring, we compare an unranked learned-prior arm with a height-informed rule controller on the same ordered 1,000-episode official TEST panel. Both use the same frozen patch-token readouts, observation-based memory, placement controller, release logic, goal frames, 225-action cap, and paper T1 success rule. The rule replaces the unranked arm's learned trajectory prior and additionally reads the non-privileged $z$ coordinate of proprioceptive end-effector position; neither arm uses predictive candidate scoring or proposer-driven candidate selection. Consequently, this is a more-informed control reference, not a same-input ablation of the learned prior. The test run follows rule design and functional validation on a disjoint 113-episode development panel.

The unranked learned-prior arm is source-sampling seed 0 of the separate Cube attribution suite, not the Source-CEM headline configuration. Neither arm uses predictive candidate scoring, and their paired difference does not estimate a difference from the headline LeCompass run.

In \textsc{PrePick}, the rule controller commands raw vertical action $-0.3$ with the gripper open (raw $-1.0$) until observed end-effector height is at most $0.026$ m. It then commands closure (raw $+1.0$) and zero vertical action; after contact is observed, it commands raw vertical action $+0.6$ until the observed end-effector height is at least $0.06$ m above its value at first contact. It maintains a closed-gripper command until downstream placement and release take ownership, and commands zero yaw throughout. The existing acquisition readout supplies $x,y$, and all later placement phases remain unchanged. These fixed task-specific rules use no privileged block position, flow decoding, or LeWM predictive evaluations.

\begin{table}[!htbp]
    \centering
    \small
    \setlength{\tabcolsep}{3pt}
    \caption{\textbf{Cube shared-stack control reference.} Counts use the same 1,000 TEST episodes; differences are height-informed rule minus unranked prior. 95\% intervals use an episode-paired normal approximation from discordant pairs without continuity correction. The rule has additional proprioceptive height. Neither arm is the headline Source-CEM configuration.}
    \label{tab:cube-simple-control}
    \begin{tabular}{@{}lrrrr@{}}
        \toprule
        Outcome & Unranked prior & Height-informed rule & Difference [95\% CI] & Exact $p$ \\
        \midrule
        Native 4-cm reach & 853 & 928 & $+7.5$ pp $[5.5,9.5]$ & $3.28\times10^{-14}$ \\
        Semantic completion & 848 & 925 & $+7.7$ pp $[5.7,9.7]$ & $1.11\times10^{-14}$ \\
        \bottomrule
    \end{tabular}
\end{table}

The T1 semantic-completion rule requires native reach, completed release, and containment at the post-action observation following the third opening command. With additional proprioceptive height and the shared execution stack, simple rules also achieve high Cube success. Both arms still use the frozen JEPA encoder through their patch-token readouts. Predictive selection without the execution stack is evaluated separately in Appendix~\ref{app:cube-bare-ranking}.

\subsection{Full-panel bounded-recovery intervention}
Table~\ref{tab:appendix-recovery-onoff} compares Recovery ON and OFF on the same 1,000-episode panel with identical frozen models, nominal configuration, and event-triggered perturbation rule; only the complete bounded recovery bundle changes. All 746 semantic-completion discordances favor ON. The 58 OFF completions also occur under ON, and all lie in the respective uninduced sets.

ON can recover from natural contact loss before the forced perturbation, so trajectories need not remain step-wise matched. The full-panel paired comparison estimates the bundle's effect; the matched-event subset below describes execution mechanisms.

\begin{table}[!htbp]
    \centering
    \small
    \caption{\textbf{Cube T3 recovery ON/OFF.} Outcomes use all 1,000 paired episodes. Differences are ON minus OFF.}
    \label{tab:appendix-recovery-onoff}
    \resizebox{\columnwidth}{!}{%
    \begin{tabular}{@{}lrrrr@{}}
        \toprule
        Outcome & Recovery ON & Recovery OFF & Difference & $p$ \\
        \midrule
        Semantic completion & 804/1000 (80.4\%) & 58/1000 (5.8\%) & +74.6 pp & $5.4\times10^{-225}$ \\
        Native 4-cm reach & 837/1000 (83.7\%) & 397/1000 (39.7\%) & +44.0 pp & $7.04\times10^{-133}$ \\
        \bottomrule
    \end{tabular}
    }
\end{table}

Semantic completion is primary because 326 induced episodes first reach the 4-cm target before sustained contact loss is confirmed in both arms. A further 450 ON and 10 OFF episodes first reach at or after loss confirmation; the remaining 61 native reaches are identical uninduced episodes. The recovery-specific reach diagnostic below instead requires target attainment after re-entry into \textsc{Carry} following the induced loss, regardless of earlier target entry.

\subsection{Matched-event recovery and additional diagnostics}
Of 839 ON episodes with induced sustained gripper-contact loss, 837 have an OFF counterpart with matching perturbation and loss-confirmation steps. Table~\ref{tab:appendix-recovery-stages} separates re-acquisition, recovery-specific reach, and completion in this matched-event subset.

\begin{table}[!htbp]
    \centering
    \small
    \caption{\textbf{Recovery on 837 matched-event episodes.} Both arms have identical perturbation and loss-confirmation steps. Recovery-specific reach requires post-loss re-entry into \textsc{Carry}; semantic completion uses the usual post-release criterion.}
    \label{tab:appendix-recovery-stages}
    \begin{tabular}{@{}lrr@{}}
        \toprule
        Post-loss outcome & Recovery ON & Recovery OFF \\
        \midrule
        Re-entry into \textsc{Carry} & 803/837 & 12/837 \\
        Recovery-specific native reach & 760/837 & 0/837 \\
        Semantic completion & 746/837 (89.1\%) & 0/837 \\
        \bottomrule
    \end{tabular}
\end{table}

Among all 839 Recovery-ON episodes with induced loss, 833 establish a new contact segment; the three subsequent outcome counts are unchanged from Table~\ref{tab:appendix-recovery-stages}. A separate physical-drop predicate using drop records and a 20-mm lift threshold identifies 678 episodes, of which 620 complete (91.4\%), distinguishing induced contact loss from evaluator-confirmed physical drops.

Recovery OFF induces sustained gripper-contact loss in 843 episodes: 825 have no subsequent contact, 14 re-grasp but do not satisfy the recovery-specific reach criterion, four release without containment after the third opening command, and none completes. Of the 14 broader re-grasps, 12 re-enter \textsc{Carry} after the induced loss without rollback, recovery assistance, or recovery budget; none completes.

\FloatBarrier
\section{Qualitative Rollouts}
\label{sec:appendix-qualitative}

\begin{figure*}[!t]
    \centering
    \includegraphics[width=\textwidth]{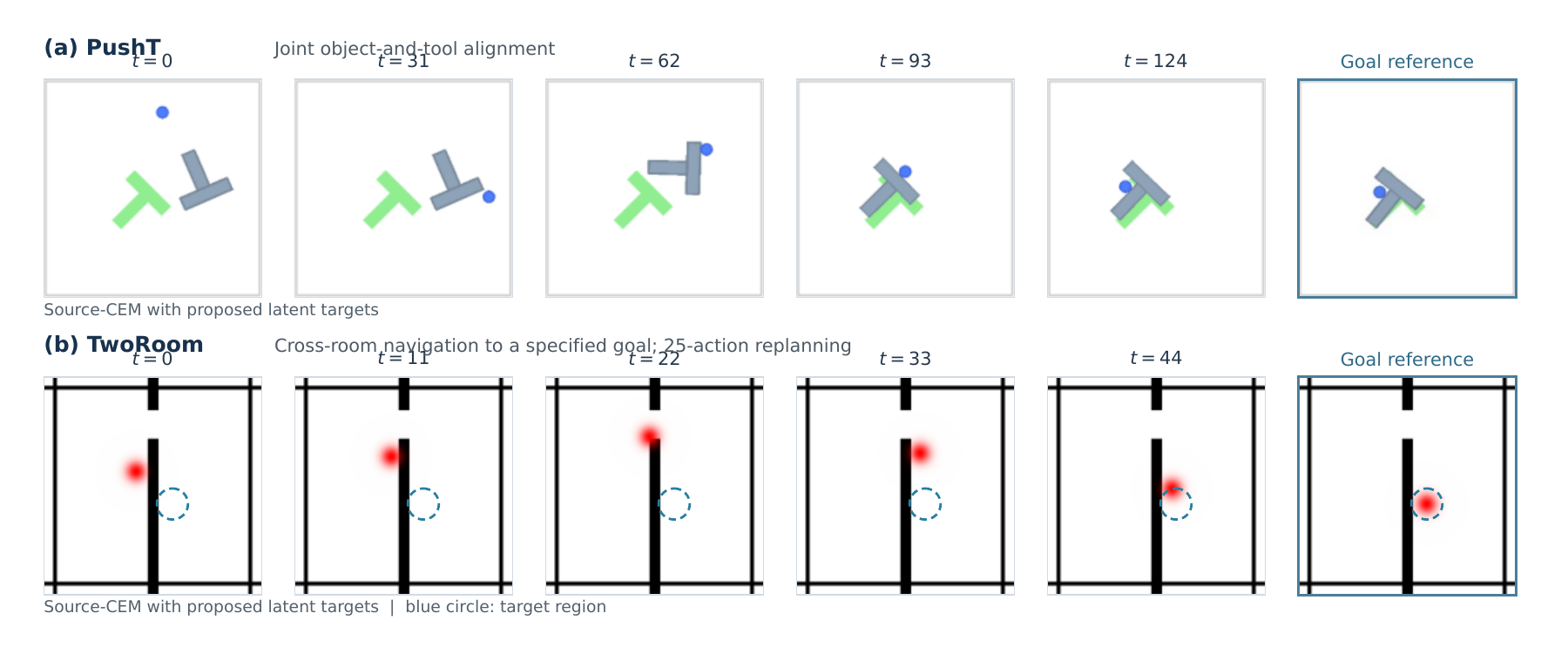}
    \caption{\textbf{Successful execution on PushT and TwoRoom.}
    (a) A pusher moves a T-shaped block to the target pose while also matching the target pusher position.
    (b) A disk-shaped agent crosses the doorway and reaches a specified position in the other room under collision-aware dynamics; the dashed blue circle marks the 16-pixel target region.
    Both sequences use CompassFlow Source-CEM with proposed latent targets, trajectory-field checkpoint 0, and 192 predictive evaluations per plan; replanning occurs every five actions on PushT and every 25 on TwoRoom.
    The five execution frames are evenly spaced between initialization and the first successful post-action observation; $t$ counts executed actions, and the rightmost column shows the goal reference.
    These are illustrative reruns: PushT uses a non-TEST case and TwoRoom a strict TEST case; neither contributes to the reported rates.}
    \label{fig:pusht-tworoom-success}
\end{figure*}

Figure~\ref{fig:pusht-tworoom-success} complements the quantitative results in Sec.~\ref{sec:results} with qualitative rollout sequences.

\newpage

\section{Additional Controlled Comparisons and Computational Cost}
\label{app:revision-evidence}

\subsection{Matched PushT search at budget 192}
\label{app:revision-coordinate}
The budget-192 comparisons share the ordered TEST500, frozen LeWM/proposer, observation history, action mapping and success predicate. All predict 25 actions, execute five-action prefixes, refresh targets every five actions and disable warm starts. Action-CEM executes its final elite mean; Source-CEM selects the lowest-cost candidate across rounds. Thus this comparison matches execution conditions but retains different priors and search rules.

\begin{table}[!htbp]\centering\small
\caption{\textbf{Paired controlled-search effects.} Differences are Source-CEM minus the named arm (percentage points); 95\% intervals use 10,000 episode-paired bootstrap draws.}
\begin{tabular}{@{}llrrl@{}}\toprule
Reference & Checkpoint & Difference & 95\% interval & $p$\\\midrule
Action-CEM & 0 & +36.6 & [31.8,41.4] & $8.48\times 10^{-40}$\\
Action-CEM & 1 & +38.6 & [33.6,43.8] & $1.66\times 10^{-40}$\\
Action-CEM & 2 & +33.6 & [28.6,38.4] & $1.49\times 10^{-34}$\\
\bottomrule
\end{tabular}\end{table}

Absolute success rates and scoring-target contrasts are reported in Table~\ref{tab:executor-results} and Appendix~\ref{sec:appendix-executor-attribution}.

\paragraph{Shared-initialization coordinate comparison.}
To isolate coordinate updates, both methods share the same initial 64 Flow-decoded trajectories, three 64-candidate rounds, eight elites, innovation seeds and global-best selection. Both fit diagonal Gaussians using population standard deviation with a $10^{-6}$ floor, insert the fitted mean as candidate zero in adaptive rounds, exclude padding and disable anchors, trust bounds, extra clamps and smoothing. Only the fitted coordinates differ: source updates are decoded through Flow, whereas action updates operate directly on normalized actions. Execution settings remain as above.

\begin{table}[!htbp]
\centering\small
\setlength{\tabcolsep}{3pt}
\caption{\textbf{Shared-initialization PushT coordinate comparison.} Each checkpoint uses the same 500 episodes. Differences are source-space minus action-space refinement in percentage points; intervals use 10,000 task-paired bootstrap draws. The mean row reports success mean $\pm$ sample SD over the three fixed checkpoints; its interval resamples paired tasks jointly across checkpoints, not 1,500 independent episodes.}
\label{tab:coordinate-matched}
\begin{tabular}{@{}lrrrrr@{}}
\toprule
Checkpoint & Source & Action & Difference & 95\% interval & $p$ \\
\midrule
0 & 277/500 (55.4\%) & 236/500 (47.2\%) & +8.2 & [3.2,13.2] & .00165 \\
1 & 263/500 (52.6\%) & 254/500 (50.8\%) & +1.8 & [-3.2,6.8] & .531 \\
2 & 254/500 (50.8\%) & 250/500 (50.0\%) & +0.8 & [-4.2,5.8] & .814 \\
\midrule
Mean & $52.93\pm2.32$\% & $49.33\pm1.89$\% & +3.6 & [0.8,6.4] & -- \\
\bottomrule
\end{tabular}
\end{table}

The mean gain is conditional on three frozen checkpoints; individual intervals exclude zero only for checkpoint 0. Equal prediction budgets do not imply equal decoding cost or latency.

\subsection{Five-budget sweep}
\label{app:revision-budget}
The sweep fixes Flow checkpoint 0 on each domain's TEST500 and varies $B\in\{48,96,192,384,768\}$. Both CEM methods use three $B/3$-candidate rounds and $B/24$ elites; Source-CEM retains its anchors and trust bounds, and Flow decoding uses eight Euler steps. PushT uses the five-action protocol above. TwoRoom executes and refreshes targets every 25 actions, updates history every five actions and retains its final-goal tail.

\begin{table}[!htbp]\centering\small
\caption{\textbf{Budget sensitivity.} Percent success on each domain's TEST 500; single fixed flow checkpoint. The TwoRoom 192 sweep run (37.0/77.4/76.0\%) is separate from the historical three-checkpoint references in Table~\ref{tab:executor-results}.}
\begin{tabular}{@{}rrrrrrr@{}}\toprule
&\multicolumn{3}{c}{PushT}&\multicolumn{3}{c}{TwoRoom}\\
$B$ & Action & IID & Source & Action & IID & Source\\\midrule
48&12.2&35.0&38.4&23.4&64.4&62.2\\
96&16.0&41.8&46.4&26.6&73.8&73.4\\
192&16.2&45.6&52.8&37.0&77.4&76.0\\
384&18.2&43.6&54.8&40.8&80.0&80.0\\
768&13.6&51.8&52.8&49.4&84.2&84.4\\\bottomrule
\end{tabular}\end{table}

Source-minus-IID paired 95\% intervals exclude zero on PushT at 192 (+7.2 pp, [2.0,12.4]) and 384 (+11.2 pp, [6.2,16.2]), but include zero at other budgets and all five TwoRoom budgets. Closed-loop success need not improve monotonically with $B$, and replanning counts can differ.

\subsection{Inference profiling and production runtime}
\label{app:revision-runtime}
Fixed-context timing uses 600 samples. PushT module statistics use non-TEST replay; TwoRoom uses online profiling. Module means are exclusive CUDA-synchronized wall times, excluding simulator stepping. Production rollout timing instead synchronizes only at rollout boundaries and excludes model loading and task setup.

\begin{table}[!htbp]\centering\small
\caption{\textbf{Budget 192 module means and decision medians (ms).} Bridge and search columns include the remaining scoring and interface work. Means across modules need not sum to the decision median.}
\begin{tabular}{@{}llrrrrrrr@{}}\toprule
Domain & Method & Encode & Prop. & Flow & WM & Bridge & Search & Median\\\midrule
PushT & Action &4.98&3.13&0&46.79&2.44&1.60&58.22\\
& IID &4.99&3.14&3.43&15.88&2.28&1.02&30.08\\
& Source &4.94&3.09&10.23&46.33&2.43&2.46&69.17\\
TwoRoom & Action &4.32&1.67&0&48.11&0.96&2.92&58.51\\
& IID &4.34&2.07&3.41&15.84&.62&2.42&29.26\\
& Source &4.50&2.13&10.57&47.84&1.00&3.91&70.01\\\bottomrule
\end{tabular}\end{table}

Source-CEM's three refinement rounds are sequential. Their Flow decoding costs about 10\,ms at budget 192, versus 3.4\,ms for one-batch IID sampling.

Production timing evaluates Direct9000, Fixed25-9000 and full LeCompass192 on 50 non-TEST tasks per domain, retaining each system's deployment configuration. PushT uses five-environment auto-reset batches, with per-task costs amortized across each batch. TwoRoom timing retains the original stop radius of 8. Five fixed tasks per domain are repeated in two additional time blocks separated by at least 900 seconds, with method order rotated across repeat blocks. Timing uses one Flow checkpoint on a shared host and a separate panel from formal success evaluation.

\begin{table}[!htbp]\centering\small
\caption{\textbf{Production execution counts and action costs.} Each triple is Direct (9000) / Fixed-25 (9000) / LeCompass (192) across each domain's 50-task panel.}
\begin{tabular}{@{}lrrr@{}}\toprule
Domain & Plans & Actions & ms / actual action\\\midrule
PushT &350 / 912 / 2126&7900 / 7900 / 7900&27.01 / 66.42 / 23.24\\
TwoRoom &282 / 265 / 198&7018 / 6491 / 4631&25.38 / 25.93 / 4.84\\
Cube &450 / 450 / 800&11250 / 11250 / 3906&194.50 / 188.08 / 209.71\\\bottomrule
\end{tabular}\end{table}

Using 10,000 paired bootstrap resamples (batches for PushT), the 95\% intervals for speedup over Direct are $[1.09,1.23]\times$, $[7.15,8.88]\times$, and $[2.25,3.18]\times$ on PushT, TwoRoom, and Cube, respectively; Source also remains faster than Direct and Fixed-25 in both cross-time repeats, although absolute runtimes vary.

\subsection{Cube execution-component study}
\label{app:revision-cube}
Table~\ref{tab:component500}(b) uses the same 500 tasks, starts and goals, frozen learned components, Source-CEM192, prefix 5, noise seed 7, reset seed 0 and cap 225. Planner-only and contact-memory-only follow the environment's stopping rule; the full interface terminates after its release check. We compare latched ever-native reach, including initial successes, rather than fixed-horizon final states or latency.

Contact-memory-only removes spatial readouts and retains contact-driven grip protection: K1=3, K2=12, loss window 3, up to three 60-step assist activations, one carry activation and no recovery. Only the grip-floor actuator can change planned actions; position and yaw pass through. Without geometry, placement confirmation and active release are unavailable.

The full interface adds acquisition/placement readouts and the execution logic that consumes them, so this comparison measures their combined benefit.

\subsection{Cube predictive ranking without the execution interface}
\label{app:cube-bare-ranking}

Both arms use the same 500 tasks as Table~\ref{tab:component500}(b), frozen encoder/proposer/Flow checkpoint 0, action mapping, noise seed 7 and reset seed 0. Each decodes 192 IID 25-action candidates and executes five-action prefixes. The ranked arm selects the first minimum-cost candidate under frozen-LeWM scoring; the unranked arm executes candidate zero without predictive scoring. Both retain the encoder and proposer.

No readouts, execution memory, channel corrections, grip protection, release program or recovery are used; all action channels pass directly to the environment. Both retain the 225-action cap and physical stopping criteria. Initial candidate clouds match; subsequent closed-loop histories may diverge.

\begin{table}[!htbp]
\centering\small
\caption{\textbf{Cube bare-planner predictive ranking.} Counts use the same 500 tasks. Native reach includes the initial frame; the excluding-initial column removes the 14 initially satisfied tasks from both numerator and denominator. True pick requires contact $>0.5$ and lift $\geq5$ cm simultaneously; carried-in requires sustained contact and lift until a new target entry under the frozen physical criterion. Drops count tasks with at least one recorded drop.}
\label{tab:cube-bare-ranking-physical}
\begin{tabular}{@{}lrrrrr@{}}
\toprule
Selection & Native & Excluding initial & True pick & Carried-in & Drops \\
\midrule
First candidate & 33/500 & 19/486 & 264 & 8 & 258 \\
WM-ranked & 127/500 & 113/486 & 227 & 75 & 166 \\
\bottomrule
\end{tabular}
\end{table}
\FloatBarrier

Predictive ranking improves native reach by 18.8 percentage points (95\% task-paired bootstrap CI $[15.0,22.6]$, 10,000 resamples; $p=2.31\times10^{-21}$), with 103 ranked-only and nine unranked-only successes. With all action channels passed directly to the environment, this controlled comparison demonstrates that frozen-LeWM predictive scoring improves closed-loop action selection without spatial readouts or an execution program. The 27.4\% Source-CEM reference uses a different search configuration and is not the ranked-IID arm. The complete-interface ranking comparison is reported in Appendix~\ref{sec:appendix-executor-attribution}.

\subsection{Terminology}
\begin{table}[!htbp]\centering\small
\caption{\textbf{Operational glossary.} Definitions refer to the evaluated implementation.}
\begin{tabular}{@{}p{.24\textwidth}p{.70\textwidth}@{}}\toprule
Term & Operational meaning\\\midrule
LeCompass & Task-level execution framework combining state-indexed targets, local frozen-JEPA prediction, trajectory search and observation-grounded control.\\
State-indexed proposer / reference & Learned proposer that recomputes a local latent reference from current observation history and the final goal, rather than following an elapsed-time waypoint schedule.\\
CompassFlow & Planning component combining a state-conditioned, target-independent Flow trajectory prior, frozen-LeWM scoring and source-space search.\\
Source & Initial noise coordinates decoded by the Flow trajectory prior into candidate action sequences.\\
Source-CEM & CEM optimization of source coordinates using frozen-LeWM scores of decoded trajectories, selecting the lowest-cost candidate across rounds.\\
Ranked IID & Independent samples from the same prior, ranked by latent distance between WM-predicted outcomes and the scoring reference, without adaptive updates.\\
Spatial readout & A supervised head reading frozen visual patch tokens to estimate execution geometry.\\
Execution memory & Explicit non-learned state updated from accumulated contact and readout evidence.\\
Action realization & Task-specific mapping from planned actions, readout estimates and execution state to environment commands.\\
Bounded recovery & Observation-triggered reacquisition after sustained contact loss, with limited retry/step budgets and retained observation history and outer-loop task progress.\\\bottomrule
\end{tabular}\end{table}

\end{document}